\documentclass[jair,twoside,11pt,theapa]{article}
\usepackage{jair, theapa, rawfonts}
\usepackage{amsmath}
\usepackage{amsthm}
\usepackage{amssymb}
\usepackage{calrsfs}
\usepackage{latexsym}
\usepackage{mathbbol} %has more characters than bbold
\usepackage[mathcal]{euscript}
\usepackage[pointlessenum]{paralist}
\usepackage[scr]{rsfso}
\usepackage{tikz}
\usetikzlibrary{cd}
\usetikzlibrary{arrows}
\usepackage{venndiagram}

\theoremstyle{definition} %title and number in bold, text normal.
\newtheorem{Defn}{Definition}[section]
\newtheorem{Refine}[Defn]{Refinement}
\newtheorem{Lem}[Defn]{Lemma}
\newtheorem{Thm}[Defn]{Theorem}
\newtheorem{Eg}[Defn]{Example}
\newtheorem{Prin}[Defn]{Principle}
\newcommand{\defn}{\textsl} %definition
\newcommand{\nl}{\newline}
\newcommand{\lb}{\linebreak}

\newcommand{\strArr}{\ensuremath{\rightarrow}}   %strict arrow
\newcommand{\defArr}{\ensuremath{\Rightarrow}} %defeasible arrow
\newcommand{\warnArr}{\ensuremath{\leadsto}}    %warning arrow

\newcommand{\AND}{\ensuremath{\mathord{\raisebox{0.32ex}{\ensuremath{\scriptstyle \bigwedge}}}}} %unary conjunction
\newcommand{\bbt}{\ensuremath{\mathbb{t}}} %a tautology \tt pre-defined
\newcommand{\calD}{\ensuremath{\mathcal{D}}} %a plausible-description
\newcommand{\calL}{\ensuremath{\mathcal{L}}} %first-order language
\newcommand{\curlyD}{\ensuremath{\mathscr{D\!}}} %a plausible-description
\newcommand{\curlyL}{\ensuremath{\mathscr{L\!}}} %a logic
\newcommand{\curlyR}{\ensuremath{\mathscr{R\!}}} %plausible-representation
\newcommand{\e}{\ensuremath{\!\in\!}} %is an element of 
\newcommand{\F}{\ensuremath{\textsf{\textbf{F}}}} %the false truth value
\newcommand{\imps}{\ensuremath{\mathrel{|}\mathrel{\mkern-3.5mu}\Relbar}} %implies f |=
\newcommand{\nimps}{\ensuremath{\mathrel{|}\mathrel{\mkern-3.5mu}\not\Relbar}} 
\newcommand{\NN}{\ensuremath{\mathbb{N}}} %Natural numbers
\newcommand{\nnot}{\ensuremath{\mathord{\sim}}} %normalised not = complement
\newcommand{\nproves}{\ensuremath{\mathrel{|}\mathrel{\mkern-4mu}\not\relbar}} 
\newcommand{\nte}{\ensuremath{\!\notin\!}} %\ne is pre-defined as \not =
\newcommand{\OR}{\ensuremath{\mathord{\raisebox{0.32ex}{\ensuremath{\scriptstyle \bigvee}}}}} %unary disjunction 
\newcommand{\proves}{\ensuremath{\mathrel{|}\mathrel{\mkern-4mu}\relbar}} %|- 
\newcommand{\T}{\ensuremath{\textsf{\textbf{T}}}} %the true truth value (tv)
\newcommand{\tv}[1]{\ensuremath{\textsf{\textbf{#1}}}} %usage \tv{t}
\newcommand{\ZZ}{\ensuremath{\mathbb{Z}}} %Integers

\newcommand{\Alg}{\ensuremath{\mathit{Alg}}} %set of all proof algorithms
\newcommand{\Ax}{\ensuremath{\mathit{Ax}}} %Axioms
\newcommand{\Ch}{\ensuremath{\mathit{Ch}}} %Ch children
\newcommand{\Cl}{\ensuremath{\mathit{Cl}}} %Cl(.) clauses
\newcommand{\Const}{\ensuremath{\mathit{Const}}} %set of constant symbols
\newcommand{\Dftd}{\ensuremath{\mathit{Dftd}}} %Defeated
\newcommand{\Doc}{\ensuremath{\mathit{Doc}}} %Domain of change
\newcommand{\Dom}{\ensuremath{\mathit{Dom}}} %Domain
\newcommand{\Fact}{\ensuremath{\mathit{Fact}}} %factual part
\newcommand{\Fml}{\ensuremath{\mathit{Fml}}} %set of all formulas
\newcommand{\Foe}{\ensuremath{\mathit{Foe}}} %evidence against
\newcommand{\For}{\ensuremath{\mathit{For}}} %evidence for
\newcommand{\Func}{\ensuremath{\mathit{Func}}} %set of function symbols
\newcommand{\Plaus}{\ensuremath{\mathit{Plaus}}} %plausible part
\newcommand{\Pred}{\ensuremath{\mathit{Pred}}} %set of predicate symbols
\newcommand{\Rul}{\ensuremath{\mathit{Rul}}} %converts clauses to strict rules
\newcommand{\Taut}{\ensuremath{\mathit{Taut}}} %set of all tautologies
\newcommand{\Thms}{\ensuremath{\mathit{Thm}}} %theorems of [\Thm is pre-defined]
\newcommand{\Var}{\ensuremath{\mathit{Var}}} %set of all variables
\ShortHeadings{Plausible Reasoning and First-Order Plausible Logic}{Billington}
\begin{document}

\title{Plausible Reasoning and First-Order Plausible Logic}

\author{\name David Billington 
\email d.billington@griffith.edu.au d.billington50@gmail.com\\
\addr School of Information and Communication Technology, Nathan campus,\\
      Griffith University, Brisbane, Queensland 4111, Australia.}

% For research notes, remove the comment character in the line below.
% \researchnote

\maketitle

\begin{abstract} 
Defeasible statements are statements that are likely, or probable, or usually true, 
but may occasionally be false. 
Plausible reasoning makes conclusions from statements that are either facts or 
defeasible statements without using numbers. 
So there are no probabilities or suchlike involved. 
Seventeen principles of logics that do plausible reasoning are suggested and 
several important plausible reasoning examples are considered. 
There are 14 necessary principles and 3 desirable principles, one of which is not formally stated. 
A first-order logic, called Plausible Logic (PL), is defined that satisfies all but two of the 
desirable principles and reasons correctly with all the examples. 
As far as we are aware, this is the only such logic. 
PL has 8 reasoning algorithms because, from a given plausible reasoning situation, 
there are different sensible conclusions. 
This article is a condensation of my book `Plausible Reasoning and Plausible Logic' (PRPL), 
which is to be submitted. 
Each section of this article corresponds to a chapter in PRPL, and vice versa. 
The proofs of all the results are in PRPL, so they are omitted in this article. 
\end{abstract}

%2345678901234567890123456789012345678901234567890123456789012345678901234567890
%Section 1 Preliminaries
\section{Preliminaries}
\label{Section:Preliminaries}
This article is a condensation of my book `Plausible Reasoning and Plausible Logic' (PRPL), 
which is to be submitted. 
Each section of this article corresponds to a chapter in PRPL, and vice versa. 
The proofs of all the results are in PRPL, so they are omitted in this article. 

The number of elements in, or cardinality of, a set $S$ is denoted by $|S|$. 
The set of all integers is denoted by $\ZZ$, and 
the set of all positive integers is denoted by $\ZZ^+$. 
The set $\NN$ of \defn{natural numbers} is defined by \,$\NN = \{0\} \cup \ZZ^+$. 
The ordinal $\omega$ is defined by \,$\omega = \NN$. 
If $m$ and $n$ are integers then we define \,$[m..n] = \{i \e \ZZ : m \leq i \leq n\}$. 
Suppose $S$ and $S'$ are sequences. 
If $S$ is finite then the concatenation of $S$ onto the left of $S'$ is denoted by \,$S$++$S'$. 
We define \,$e$+$S = (e)$++$S$, and if $S$ is finite then we define \,$S$+$e = S$++$(e)$. 

A \defn{directed graph}, also called a \defn{digraph}, is a graph in which 
every edge has a direction. 
The common term directed acyclic graph (dag) is an acyclic directed graph, 
rather than an acyclic graph with directed edges as the name suggests. 
Also dag has insulting and unpleasant connotations. 
An acyclic digraph is a representation of an acyclic binary relation. 
Since we regard a rooted acyclic digraph (rad) as a generalisation of a rooted directed tree, 
we shall use tree nomenclature where appropriate. 
A rooted acyclic digraph (rad) allows a node to have more than one parent. 
The definition and notation that we shall use follows. 

%Definition 1.1
\begin{Defn} \label{Defn:rad} 
Suppose $N$ is a non-empty countable set of \defn{nodes} (also called points), 
$A$ is an acyclic binary relation on $N$, the elements of $A$ are called \defn{arcs}, and 
\,$r \e N$\, is called the \defn{root} node. 
Let \,$T = (N,A,r)$\, and consider the following conditions that $T$ may satisfy. 
\begin{compactenum}[1)]
\item \label{r has no parents} %1)
	For all $n$ in $N$, \,$(n,r) \nte A$. 
\item \label{root to node path} %2)
	For all $n$ in $N\!-\!\{r\}$, there exists $k$ in $\ZZ^+$ such that 
	for all $i$ in $[0..k\!-\!1]$, \,$(n_i,n_{i+1}) \e A$, 
	where \,$n_0 = r$\, and \,$n_k = n$. 
\item \label{levels} %3) This creates levels. 
	For all $n$ in $N\!-\!\{r\}$, there exists exactly one $l$ in $\ZZ^+$ such that 
	for all $i$ in $[0..l\!-\!1]$, \,$(n_i,n_{i+1}) \e A$, 
	where \,$n_0 = r$\, and \,$n_l = n$. 
\item \label{parents are unique} %4) This makes a rad into a tree. 
	For all $c$ in $N\!-\!\{r\}$ there is exactly one $p$ in $N$ such that \,$(p,c) \e A$. 
\end{compactenum}
$T$ is a \defn{rooted acyclic digraph} (\defn{rad}) iff both 
(\ref{r has no parents}) and (\ref{root to node path}) hold. \nl
$T$ is a \defn{layered rooted acyclic digraph} iff both 
(\ref{r has no parents}) and (\ref{levels}) hold. \nl
$T$ is a \defn{rooted directed tree} iff 
(\ref{r has no parents}), (\ref{levels}), and (\ref{parents are unique}) all hold. 
\end{Defn} 

In Definition~\ref{Defn:rad}, (\ref{levels}) implies (\ref{root to node path}), 
so a layered rooted acyclic digraph is a rooted acyclic digraph, and 
a rooted directed tree is a layered rooted acyclic digraph and hence a rooted acyclic digraph. 

Let \,$T = (N,A,r)$\, be a rooted acyclic digraph (rad). 
We say $n$ is a node of, or in, $T$ iff \,$n \e N$. 
Define $|T|$ to be the number of nodes in $T$, so \,$|T| = |N|$. 
We say \defn{$T$ is finite} iff $|T|$ is finite. 
If $(p,c) \e A$ then $p$ is called a \defn{parent} of $c$ and 
$c$ is called a \defn{child} of $p$. 
Define \,$\Ch(T,p) = \{c \e N : (p,c) \e A\}$\, to be the set of all children in $T$ of the node $p$. 
If $T$ is clear from the context we often write $\Ch(p)$ instead of $\Ch(T,p)$. 
A node is called a \defn{leaf} iff it has no children. 
A \defn{path} in $T$ from $p_1 \e N$ to $p_k \e N$ is a sequence of 
$k$ different nodes $(p_1, p_2, ..., p_k)$ such that \,$k \geq 2$\, and 
for each $i$ in $[1..k\!-\!1]$, $(p_i,p_{i+1})$ is an arc of $T$. 
The (usual definition of the) \defn{length} of the path $(p_1, p_2, ..., p_k)$ is $k\!-\!1$. 
(This is inconsistent with the (usual) definition of the length of a sequence.) 
An \defn{infinite path} in $T$ is a sequence $(p_1, p_2, ...)$ of infinitely many 
nodes of $T$ such that for each $i$ in $\ZZ^+$, $(p_i,p_{i+1})$ is an arc of $T$. 
The \defn{length} of an infinite path is $\omega$. 
The set of \defn{ancestors} of $c \e N$ in $T$ is 
\,$\{n \e N :$ there is a path in $T$ from $n$ to $c\}$. 
The set of \defn{descendants} of $p \e N$ in $T$ is 
\,$\{n \e N :$ there is a path in $T$ from $p$ to $n\}$. 
A rad \,$T' = (N',A',r')$\, is a \defn{subrad} of $T$ iff \,$N' \subseteq N$\, and \,$A' \subseteq A$. 

The following lemma gives a characterisation of when a rooted acyclic digraph is finite. 

%%%%%%%%%%%%%%%%%%%%%%%%%%%%%%%%%%%%%%%%%%
%Lemma~1.2
\begin{Lem} 
\label{Lem:rad is finite iff ...} 
Let $T$ be a rooted acyclic digraph. 
$T$ is finite iff 
\begin{compactenum}[1)]
\item  %1)
	each node in $T$ has only finitely many children, and 
\item  %2)
	each path in $T$ has only finitely many nodes. 
\end{compactenum}
\end{Lem} %Lemma~1.2
%%%%%%%%%%%%%%%%%%%%%%%%%%%%%%%%%%%%%%%%%%

The following theorem allows induction on rooted acyclic digraphs (rads). 

%%%%%%%%%%%%%%%%%%%%%%%%%%%%%%%%%%%%%%%%%%
%Theorem~1.3
\begin{Thm}  [rad induction] 
\label{Thm:rad induction} 
Let \,$T = (N,A,r)$\, be a rooted acyclic digraph (rad). 
For each node $p$ in $N$, let $S(p)$ be a statement about $p$ which is either true or false, 
but not both. 
Suppose that (1) and (2) both hold. 
\begin{compactenum}[1)]
\item  %1)
	If \,$p \e N$\, and for all $c$ in $\Ch(p)$, $S(c)$ is true; then $S(p)$ is true. \nl 
	(So if $l$ is a leaf then $S(l)$ is true.) 
\item  %2)
	Each path in $T$ has only finitely many nodes. 
\end{compactenum}
Then for all $p$ in $N$, $S(p)$ is true. 
\end{Thm} %Theorem~1.3
%2345678901234567890123456789012345678901234567890123456789012345678901234567890

%2345678901234567890123456789012345678901234567890123456789012345678901234567890
%Section 2
\section{First-Order Resolution Logic (FORL)}
\label{Section:FORL}
First-order classical logic underlies first-order Plausible Logic. 
The inference mechanism for our classical logic (FORL) is resolution. 
Let $f$ be a formula and $F$ be a finite (possibly empty) set of formulas. 
Then $\neg f$ denotes the negation of $f$, $\AND F$ denotes the conjunction of $F$, and 
$\OR F$ denotes the disjunction of $F$. 
A \defn{tautology} (or valid formula) is any formula that is satisfied by every model. 
$\Taut$ is the set of all tautologies. 
$\Fml$ is the set of all formulas. 

%2345678901234567890123456789012345678901234567890123456789012345678901234567890
%Section 3
\section{Principles of Logics for Plausible Reasoning} 
\label{Section:Principles}
We are interested in reasoning about situations that 
(a) have imprecisely defined parts, and 
(b) this lack of precision is not quantified. 
Point (b) means that there are no numbers like probabilities, 
that would quantify the lack of precision. 
These situations are often indicated by the ordinary, rather than technical, 
use of words such as `mostly', `usually', `typically', `normally', `probably', 
`likely', `plausible', `believable', and `reasonable'. 
Although these words are not synonymous, they share a common property, 
which may be expressed by using either frequency of occurrence or weight of evidence. 
In frequency terms the property is that something is true more often than not; 
in evidence terms the property is that 
the evidence for something outweighs the evidence against it. 
Two examples of the kind of situations we have in mind are: `Most mammals have fur' and 
`After a roll of a fair standard die the uppermost side is usually not 1'. 
We shall call these situations \textit{plausible-reasoning situations} 
and the reasoning used in such situations \textit{plausible reasoning}. 

How do you determine whether a logic does plausible reasoning or not? 
One way to start to answer the question is to make a list of plausible reasoning examples. 
If a logic does not get the right answer to all of these examples then 
clearly the logic does not do plausible reasoning. 
So these examples act as counter-examples to show that 
a given logic does not do plausible reasoning. 
Like all good counter-examples they should be as simple as possible. 

Another role that a plausible reasoning example can play is to be a signpost to 
general principles that logics that do plausible reasoning should satisfy. 
This section introduces 5 such signpost examples and 17 principles that give a 
much clearer understanding of what it means for a formal logic to do plausible reasoning. 
There are 14 necessary principles and 3 desirable principles, one of which is not formally stated. 
If a logic fails a necessary principle then it does not do plausible reasoning. 
If a logic fails a desirable principle then it might be regarded as doing plausible reasoning but 
with limitations. 

The hope is that this set of principles will characterise 
the formal logics that do plausible reasoning. 
Whether it does or not, this seems to be the first such set of principles.
Earlier versions of these principles are in \cite{Billington2017,Billington2019,Billington2024}, 
one of these principles has been replaced and another has been weakened, 
requiring a new principle. 
However, on page 114 of \cite{WTG2014} there is a list of 11 characteristics of plausible reasoning, 
rather than characteristics of formal logics that do plausible reasoning. 

Lists of postulates, properties, or principles that concern special types of reasoning are useful 
for at least the following reasons. 
\begin{compactenum}[1)] %\raggedright
\item They help \textit{characterise} the intended special type of reasoning. 
\item They provide a means of \textit{evaluating} existing reasoning systems to see 
	how well they perform the intended special type of reasoning. 
\item They provide \textit{guidelines} for creating new reasoning systems for 
	the intended special type of reasoning. 
\item They explicitly show a \textit{difference} between the intended special type of reasoning 
	and an existing form of reasoning. 
\end{compactenum}
%%%%%%%%%%%%%%%%%%%%%%%%%%%%%%%%%%%%%%%%%

%Subsection 3.1
\subsection{Plausible-representations} 
\label{PLPR:Plausible-representations}
We shall only consider those plausible-reasoning situations that can be specified by a 
\defn{plausible-representation} \,$\curlyR = (\Fact(\curlyR), \Plaus(\curlyR))$, where 
$\Fact(\curlyR)$ is a satisfiable set of sentences representing the factual part of $\curlyR$, 
and $\Plaus(\curlyR)$ is a set representing the plausible part of $\curlyR$. 
The elements of $\Plaus(\curlyR)$ can have a variety of forms, and are often not even formulas. 
For example: defaults are used in Reiter's Default Logic \cite{Reiter1980}. 
The plausible-representation construct is very general, 
while being specific enough to permit the definition of concepts needed later. 
A degenerate plausible-representation, $\curlyR$, can have 
$\Plaus(\curlyR)$ being the empty set $\{\}$. 
Plausible-representations can easily be generalised to accommodate non-classic logics. 

Several examples involve an $n$-lottery based on $[1..n]$; which we now define. 

%Definition~3.1
\begin{Defn} \label{Defn:n-lottery} 
%\raggedright \parindent = 1.2em
If \,$n \e \ZZ^+$\, then an \defn{$n$-lottery} based on $[1..n]$ randomly selects 
a number from $[1..n]$. 
We shall often use $s_i$ to denote that the selected number was $i$. 
\end{Defn} 

Several equivalents to an $n$-lottery occur in the literature; 
for example an $n$-sided die, or a (2-sided) coin, or an $n$-sided top, 
or an urn containing $n$ balls appropriately numbered. 

Our first signpost example is a 3-lottery. 

%Example~3.2
\begin{Eg}[The 3-lottery example] 
\label{Eg:3-lottery}
%\raggedright \parindent = 1.2em
Consider the 3-lottery based on $[1..3]$. 
Then we have the following. 
\begin{compactenum}[1)] 
\item Exactly one element of $\{s_1, s_2, s_3\}$ is true. 
\item Each element of $\{s_1, s_2, s_3\}$ is unlikely. 
\item Each element of $\{\neg s_1, \neg s_2, \neg s_3\}$ is likely. 
\end{compactenum}
\end{Eg}

This example illustrates some important properties of plausible reasoning 
that will be considered later. 
The following notation will be convenient. 

%Definition~3.3
\begin{Defn} \label{Defn:classical, Thm(L,a,R), Thm(S)} 
%\raggedright \parindent = 1.2em
Suppose \,$Z \in \{$language, formula, sentence, logic$\}$. 
By a \defn{classical} $Z$ we shall mean either a propositional or a first-order $Z$. 
$\Thms(\curlyL,\alpha,\curlyR)$ denotes the set of all classical sentences 
that are derivable by the logic $\curlyL$'s proof algorithm $\alpha$ 
from the plausible-representation $\curlyR$. 
If $S$ is a set of classical sentences then $\Thms(S)$ denotes the set of 
all the classical sentences that are derivable from $S$ by (the proof algorithm of) a classical logic. 
(This simpler notation is unambiguous because $\Thms(S)$ is independent of 
the logic (for example Hilbert systems, or resolution systems) and its proof algorithm.) 
\end{Defn} 
%%%%%%%%%%%%%%%%%%%%%%%%%%%%%%%%%%%%%%%%%

%Subsection 3.2
\subsection{Representation} 
\label{PLPR:Representation}
Plausible-reasoning situations may contain facts as well as plausible information. 
For instance in the context of Example \ref{Eg:3-lottery}, the 3-lottery example, 
the formula $\OR\{s_1, s_2, s_3\}$ is a fact and always true; 
whereas the formula $\neg s_1$ is not a fact but only a plausible statement. 
Semantically these are very different formulas and 
so they need to be differentiated syntactically. 
Hence the following necessary principle of plausible reasoning. 

%Principle~3.4
\begin{Prin}[The Representation Principle] 
\label{Prin:Representation} 
%\raggedright \parindent = 1.2em
A logic for plausible reasoning must be able to represent, and distinguish between, 
factual and plausible statements. 
\end{Prin}

Reasoning makes conclusions from given information. 
It is sometimes useful for conclusions to contain more information than was used to make them. 
Such reasoning is called ampliative reasoning. 
Inductive reasoning and abductive reasoning are examples of ampliative reasoning. 
In particular, it is possible for a conclusion to be more precise than the information 
used to generate it; and so it contains more information than was used to make it. 
However, the kind of reasoning that we wish to characterise in these principles 
explicitly excludes ampliative reasoning. 
The following necessary principle excludes ampliative reasoning. 

%Principle~3.5
\begin{Prin}[The Information Principle] 
\label{Prin:Information} 
%\raggedright \parindent = 1.2em
A sentence proved by a logic for plausible reasoning must not contain more information than 
was used to derive it. 
\end{Prin}

We infer from Principle~\ref{Prin:Representation} and Principle~\ref{Prin:Information} that 
we should be able to distinguish between conclusions 
that are factual and those that are merely plausible. 
A simple way of making this distinction is to have a 
\defn{factual proof algorithm} that only uses facts and deduces only facts, 
and also a \defn{plausible proof algorithm} 
that may use facts and plausible statements  
and deduces sentences that are only plausible. 
We shall denote the factual proof algorithm by $\varphi$ (f for factual and phi). 
%%%%%%%%%%%%%%%%%%%%%%%%%%%%%%%%%%%%%%%%%

%Subsection 3.3
\subsection{The Factual Proof Algorithm} 
\label{PLPR:Factual Proof Algorithm}
The factual proof algorithm, $\varphi$, of a logic for plausible reasoning 
should be subclassical; that is, if a fact is proved by $\varphi$ 
then it should be provable in a classical logic; in symbols
$\Thms(\curlyL,\varphi,\curlyR) \subseteq \Thms(\Fact(\curlyR))$. 
The most serious problem with $\Thms(\Fact(\curlyR))$ is that 
if $\Fact(\curlyR)$ is unsatisfiable then it contains all sentences, even contradictions. 
We have avoided this by defining $\Fact(\curlyR)$ to be a satisfiable set of sentences. 
Even so, $\Thms(\Fact(\curlyR))$ contains sentences that 
we do not want to force logics for plausible reasoning to prove. 
Such sentences include tautologies, particularly ones that are not relevant to $\Fact(\curlyR)$. 
Moreover, suppose $f$ is a proved fact; 
that is, \,$f \e \Thms(\curlyL,\varphi,\curlyR)$, and $\bbt$ is any tautology. 
Then \,$\AND\{f, \bbt\} \in \Thms(\Fact(\curlyR))$. 
But we do not want to force $\AND\{f, \bbt\}$ to be in $\Thms(\curlyL,\varphi,\curlyR)$. 
The following necessary principle gives the minimum requirements that $\varphi$ must satisfy. 

%Principle~3.6
\begin{Prin}[The Factual Subclassicality Principle] 
\label{Prin:Factual Subclassicality} 
%\raggedright \parindent = 1.2em
Let $\curlyL$ be a logic for plausible reasoning and 
suppose $\varphi$ is the factual proof algorithm of $\curlyL$. 
Let $\curlyR$ be a plausible-representation. 
Let \,$X = \{\AND\{s, \bbt\} \in \Thms(\Fact(\curlyR)) : 
s \e \Thms(\curlyL,\varphi,\curlyR)$, and $\bbt \e \Taut\}$. 
Then \,$\Thms(\Fact(\curlyR)) - (\Taut \cup X) \,\subseteq\, 
\Thms(\curlyL,\varphi,\curlyR) \,\subseteq\, \Thms(\Fact(\curlyR))$. 
\end{Prin}

On page 9 of \cite{Billington2017}, page 134 of \cite{Billington2019}, and in 
\cite{Billington2024} there is the Plausible Superclassicality Principle which says 
`what is true is usually true'. 
This principle has been replaced by Principle~\ref{Prin:Factual Subclassicality}, 
as there is nothing wrong with a plausible proof algorithm that only proves plausible formulas. 
%%%%%%%%%%%%%%%%%%%%%%%%%%%%%%%%%%%%%%%%%

%Subsection 3.4
\subsection{Evidence and Non-Monotonicity} 
\label{PLPR:Evidence, Non-Monotonicity}
Let us now see if we can establish some general guidelines 
concerning the plausible provability of a given sentence $s$. 
A plausible-reasoning situation will have evidence for and against $s$. 
So it seems reasonable to determine whether $s$ is provable or not 
by just comparing these two sets of evidence, 
and declaring $s$ to be provable iff the preponderance of evidence is for $s$. 
This leads to the following necessary principle. 

%Principle~3.7
\begin{Prin}[The Evidence Principle] 
\label{Prin:Evidence}
%\raggedright \parindent = 1.2em
Let $\alpha$ be a plausible proof algorithm of a logic $\curlyL$, and 
let $\curlyR$ be a plausible-representation.  
Then \,$s \e \Thms(\curlyL,\alpha,\curlyR)$\, iff 
all the evidence in $\curlyR$ for $s$ sufficiently outweighs 
all the evidence in $\curlyR$ against $s$. 
\end{Prin}

Exactly what constitutes evidence for or against $s$ can only be determined 
when $\curlyL$, $\alpha$, and $\curlyR$ are known. 
Also `sufficiently outweighs' depends on the intuition that is being modelled, 
as well as $\curlyL$, $\alpha$, and $\curlyR$. 

A consequence of the evidence principle is non-monotonicity, which we now define. 

%Definition~3.8
\begin{Defn} \label{Defn:[non-]monotonic} 
%\raggedright \parindent = 1.2em
A proof algorithm $\alpha$ of a logic $\curlyL$ is said to be \defn{monotonic} iff 
for any two plausible-representations, $\curlyR_1$ and $\curlyR_2$, %\nl
if \,$\Fact(\curlyR_1) \!\subseteq\! \Fact(\curlyR_2)$\, and 
\,$\Plaus(\curlyR_1) \!\subseteq\! \Plaus(\curlyR_2)$\, then 
\,$\Thms(\curlyL,\alpha,\curlyR_1) \!\subseteq\! 
\Thms(\curlyL,\alpha,\curlyR_2)$. 
A proof algorithm is said to be \defn{non-monotonic} iff it is not monotonic. 
\end{Defn} 

Any proof algorithm of a classical logic is monotonic. 
However, a plausible proof algorithm must be non-monotonic because 
the addition of evidence against a previously provable sentence 
can cause it to be unprovable, as shown in our second signpost example. 

%Example~3.9
\begin{Eg}[The Non-Monotonicity example] 
\label{Eg:Non-Monotonicity}
%\raggedright \parindent = 1.2em
Cephalopods are marine animals that have tentacles; 
octopuses, squids, cuttlefish, and nautiluses are all cephalopods. 
Consider the following four statements. 
\begin{compactenum}[1)]
\item Nautiluses are cephalopods. 
\item Cephalopods usually do not have external shells. 
\item Nautiluses have external shells. 
\item Nancy is a cephalopod. 
\end{compactenum}
From these four statements it is reasonable to conclude that 
Nancy probably does not have an external shell. 
But suppose that later we discover the following statement. \nl
5) Nancy is a nautilus. \nl
Then it is not reasonable to conclude that Nancy probably does not have an external shell. 
Indeed these five statements lead to the conclusion that Nancy has an external shell. 
\end{Eg}

The non-monotonicity example shows the necessity of our next principle. 

%Principle~3.10
\begin{Prin}[The Non-Monotonicity Principle] 
\label{Prin:Non-Monotonicity}
%\raggedright \parindent = 1.2em
A plausible proof algorithm must be non-monotonic. 
\end{Prin}
%%%%%%%%%%%%%%%%%%%%%%%%%%%%%%%%%%%%%%%%%

%Subsection 3.5
\subsection{Conjunction} 
\label{PLPR:Conjunction}
Let us start by defining what we mean by a conjunctive proof algorithm. 

%Definition~3.11
\begin{Defn} \label{Defn:conjunctive} 
%\raggedright \parindent = 1.2em
A proof algorithm $\alpha$ of a logic $\curlyL$ is said to be \defn{conjunctive} iff 
for any plausible-representation, $\curlyR$, and any two sentences, $f$ and $g$, 
whenever we have \,$\{f,g\} \!\subseteq\! \Thms(\curlyL,\alpha,\curlyR)$\, then 
\,$\AND \{f,g\} \e \Thms(\curlyL,\alpha,\curlyR)$. 
A proof algorithm is said to be \defn{non-conjunctive} iff it is not conjunctive. 
\end{Defn} 

The proof algorithm of any classical logic is conjunctive. 

Conjunctions of plausible formulas behave very differently from 
conjunctions of formulas that are certain. 
In Example~\ref{Eg:3-lottery}, \,$\AND\{\neg s_1, \neg s_2\}$\, is equivalent to \,$s_3$. 
Hence, although $\neg s_1$ is plausible and $\neg s_2$ is plausible, 
\,$\AND\{\neg s_1, \neg s_2\}$\, is not plausible. 
Therefore plausible proof algorithms are not conjunctive. 
Hence the following principle is necessary. 

%Principle~3.12
\begin{Prin}[The Non-Conjunctive Principle] 
\label{Prin:Non-Conjunctive} 
%\raggedright \parindent = 1.2em
A plausible proof algorithm must not be conjunctive. 
\end{Prin}

Although the conjunction of two plausible formulas is not necessarily plausible, 
the conjunction of two facts is a fact. 
So what about the conjunction of a fact and a plausible formula? 
Clearly it cannot be a fact, but is it always plausible? 
Intuitively, a fact $f$ is always true, 
and a plausible formula $g$ is true more often than not. 
So \,$\AND\{f, g\}$\, should be true whenever $g$ is true, 
and hence \,$\AND\{f, g\}$\, should be plausible. 
This leads to the following definition and principle which is necessary. 

%Definition~3.13
\begin{Defn} \label{Defn:plausibly conjunctive} 
%\raggedright \parindent = 1.2em
Let $\curlyL$ be a logic for plausible reasoning, and 
$\varphi$ be its factual proof algorithm. 
A plausible proof algorithm $\alpha$ of $\curlyL$ is said to be \defn{plausibly conjunctive} iff 
for any plausible-representation, $\curlyR$, and any sentences, $f$ and $g$, 
whenever \,$f \e \Thms(\curlyL,\varphi,\curlyR)$\, and \,$g \e \Thms(\curlyL,\alpha,\curlyR)$\, 
then \,$\AND \{f,g\} \e \Thms(\curlyL,\alpha,\curlyR)$. 
\end{Defn} 

%Principle~3.14
\begin{Prin}[The Plausibly Conjunctive Principle] 
\label{Prin:Plausibly Conjunctive} 
%\raggedright \parindent = 1.2em
A plausible proof algorithm must be plausibly conjunctive. 
\end{Prin}
%%%%%%%%%%%%%%%%%%%%%%%%%%%%%%%%%%%%%%%%%

%Subsection 3.6
\subsection{Disjunction} 
\label{PLPR:Disjunction}
It could happen that whenever a disjunction is provable 
then at least one of its disjuncts is provable. 
When this happens we shall call the proof algorithm right disjunctive and 
the disjunction is often called point-wise disjunction. 
The formal definition follows. 

%Definition~3.15
\begin{Defn} \label{Defn:right disjunctive} 
%\raggedright \parindent = 1.2em
A proof algorithm $\alpha$ of a logic $\curlyL$ is said to be 
\defn{right disjunctive} iff for any plausible-representation, $\curlyR$, and 
any sentences, $f$ and $g$, 
whenever we have \,$\OR \{f,g\} \e \Thms(\curlyL,\alpha,\curlyR)$\, 
then either \,$f \e \Thms(\curlyL,\alpha,\curlyR)$\, or 
\,$g \e \Thms(\curlyL,\alpha,\curlyR)$. 
\end{Defn} 

The proof algorithm of any classical logic is not right disjunctive. 
The 3-lottery example (Example~\ref{Eg:3-lottery}) shows that, 
although $s_1$ and $s_2$ are both unlikely, their disjunction $\OR\{s_1, s_2\}$ is likely. 
Hence our next principle is necessary. 

%Principle~3.16
\begin{Prin}[The Not Right Disjunctive Principle] 
\label{Prin:Not Right Disjunctive} 
%\raggedright \parindent = 1.2em
A plausible proof algorithm must not be right disjunctive. 
\end{Prin}

In classical logic we have the following left disjunctive property. 
Let $F$ be a set of sentences, and let each of $f$, $g$, and $h$ be a sentence. 
If \,$F \!\cup\! \{g\} \imps f$\, and \,$F \!\cup\! \{h\} \imps f$\, then 
\,$F \!\cup\! \{\OR\{g,h\}\} \imps f$. 
A translation of this property using plausible-representations is given in the following definition.  

%Definition~3.17
\begin{Defn} \label{Defn:left factual disjunctive property} 
%\raggedright \parindent = 1.2em
A proof algorithm $\alpha$ of a logic $\curlyL$ satisfies the \defn{left factual disjunctive} \lb
property iff for any plausible-representation, $\curlyR$, and any sentences, $g$ and $h$, if \lb
\,$f \in \Thms(\curlyL,\alpha,(\Fact(\curlyR)\!\cup\!\{g\}, \Plaus(\curlyR)))$\, and 
\,$f \in \Thms(\curlyL,\alpha,(\Fact(\curlyR)\!\cup\!\{h\}, \Plaus(\curlyR)))$\, then 
\,$f \in \Thms(\curlyL,\alpha,(\Fact(\curlyR)\!\cup\!\{\OR\{g,h\}\}, \Plaus(\curlyR)))$. 
\end{Defn} 

To help discover whether a plausible proof algorithm should satisfy the  
left factual disjunctive property or not, consider our third signpost example. 

%Example~3.18
\begin{Eg}[The Left Factual Disjunction example] 
\label{Eg:Left Factual Disjunction}
%\raggedright \parindent = 1.2em
Let $\curlyR$ be a plausible-representation that models the 7-lottery based on $[1..7]$. \nl
Let $g$ be \,$\AND\{\neg s_1, \neg s_2\}$\, and 
\,$\curlyR\!+\!g = (\Fact(\curlyR)\!\cup\!\{g\}, \Plaus(\curlyR))$. \nl
Let $h$ be \,$\AND\{\neg s_3, \neg s_4\}$\, and 
\,$\curlyR\!+\!h = (\Fact(\curlyR)\!\cup\!\{h\}, \Plaus(\curlyR))$. \nl 
Let \,$\curlyR+\OR\{g,h\} = (\Fact(\curlyR)\!\cup\!\{\OR\{g,h\}\}, \Plaus(\curlyR))$. \nl 
Let $f$ be $\OR\{s_5, s_6, s_7\}$. \nl 
Then we have the following. 
\begin{compactenum}[1)] 
\item In $\curlyR$ exactly one element of $\{s_1, s_2, s_3, s_4, s_5, s_6, s_7\}$ is true. \nl
	Hence in $\curlyR$, $f$ is unlikely. 
\item In $\curlyR\!+\!g$ exactly one element of $\{s_3, s_4, s_5, s_6, s_7\}$ is true. \nl
	Hence in $\curlyR\!+\!g$, $f$ is likely. 
\item In $\curlyR\!+\!h$ exactly one element of $\{s_1, s_2, s_5, s_6, s_7\}$ is true. \nl
	Hence in $\curlyR\!+\!h$, $f$ is likely. 
\end{compactenum}
But \,$\Fact(\curlyR) \equiv \Fact(\curlyR)\!\cup\!\{\OR\{g,h\}\}$. 
So in \,$\curlyR+\OR\{g,h\}$\, exactly one element of \nl
\,$\{s_1, s_2, s_3, s_4, s_5, s_6, s_7\}$\, is true. 
Hence in \,$\curlyR+\OR\{g,h\}$, $f$ is unlikely. 
\end{Eg}

Example~\ref{Eg:Left Factual Disjunction} shows that 
a plausible proof algorithm must not satisfy the left factual disjunctive property. 
Hence our next principle is necessary. 

%Principle~3.19
\begin{Prin}[The Not Left Factual Disjunction Principle] 
\label{Prin:Not Left Factual Disjunction} 
%\raggedright \parindent = 1.2em
A plausible proof algorithm must not satisfy the left factual disjunctive property. 
\end{Prin}
%%%%%%%%%%%%%%%%%%%%%%%%%%%%%%%%%%%%%%%%%

%Subsection 3.7
\subsection{Right Weakening} 
\label{PLPR:Right Weakening}
Right Weakening can be thought of as closure under classical inference, 
as the following definition shows. 

%Definition~3.20
\begin{Defn} \label{Defn:right weakening property} 
%\raggedright \parindent = 1.2em
A proof algorithm $\alpha$ of a logic $\curlyL$ satisfies the \defn{right weakening property} iff 
for any plausible-representation, $\curlyR$, if \,$f \e \Thms(\curlyL,\alpha,\curlyR)$\, 
then \,$\Thms(\{f\}) \subseteq \Thms(\curlyL,\alpha,\curlyR)$. 
\end{Defn} 

Since $\Thms(\{f\})$ contains sentences that should not be forced to be in 
$\Thms(\curlyL,\alpha,\curlyR)$, closure under full classical inference is not what we want. 
Furthermore, suppose that whenever the facts of a plausible-representation 
$\curlyR$ and a sentence $f$ are true then the sentence $g$ is also true. 
Then in the situation defined by $\curlyR$, $g$ is true at least as often as $f$. 
So if $f$ is usually true then $g$ should also be usually true. 
Hence the next definition. 

%Definition~3.21
\begin{Defn} \label{Defn:plausible right weakening property} 
%\raggedright \parindent = 1.2em
Suppose that $\curlyL$ is any logic for plausible reasoning, and 
$\varphi$ is its factual proof algorithm. 
If $f$ is any sentence and $\curlyR$ is any plausible-representation then 
define \,$\curlyR\!+\!f = (\Fact(\curlyR)\!\cup\!\{f\}, \Plaus(\curlyR))$. 
A plausible proof algorithm $\alpha$ of $\curlyL$ has the \defn{plausible right weakening property} 
iff for any plausible-representation, $\curlyR$, if \,$f \e \Thms(\curlyL,\alpha,\curlyR)$\, 
then \,$\Thms(\curlyL,\varphi,\curlyR\!+\!f) \subseteq \Thms(\curlyL,\alpha,\curlyR)$. 
\end{Defn} 

If $\alpha$ does not satisfy the plausible right weakening property 
then there are sentences that should be provable but are not. 
Such an algorithm is insufficient; but it can be easily augmented so that 
the plausible right weakening property does hold. 
Hence the corresponding principle is regarded as necessary. 

%Principle~3.22
\begin{Prin}[The Plausible Right Weakening Principle] 
\label{Prin:Plausible Right Weakening} 
%\raggedright \parindent = 1.2em
Any plausible proof algorithm must satisfy the plausible right weakening property. 
\end{Prin}
%%%%%%%%%%%%%%%%%%%%%%%%%%%%%%%%%%%%%%%%%

%Subsection 3.8
\subsection{Consistency} 
\label{PLPR:Consistency}
Classical reasoning about a satisfiable set of facts always produces a satisfiable set of conclusions. 
So it is a fundamental shock that this is not true for plausible reasoning. 
First we need the following two definitions that relate consistency and the number of statements. 

%Definition~3.23
\begin{Defn} \label{Defn:n-consistent} 
%\raggedright \parindent = 1.2em
A proof algorithm $\alpha$ of a logic $\curlyL$ is \defn{$n$-consistent} iff 
for any plausible-representation, $\curlyR$, and any set, $S$, of sentences, 
if \,$S \!\subseteq\! \Thms(\curlyL,\alpha,\curlyR)$, and \,$|S| \leq n$\, then $S$ is satisfiable. 
\end{Defn} 

Recall that if $\curlyR$ is a plausible-representation then 
$\Fact(\curlyR)$ is a satisfiable set of sentences. 

%Definition~3.24
\begin{Defn} \label{Defn:strongly n-consistent} 
%\raggedright \parindent = 1.2em
A proof algorithm $\alpha$ of a logic $\curlyL$ is \defn{strongly $n$-consistent} iff 
for any plausible-representation, $\curlyR$, and any set, $S$, of sentences, 
if \,$S \!\subseteq\! \Thms(\curlyL,\alpha,\curlyR)$, and \,$|S| \leq n$\, 
then \,$\Fact(\curlyR) \!\cup\! S$\, is satisfiable. 
\end{Defn} 

If $\varphi$ is the proof algorithm of a classical logic $\curlyL$ then $\Plaus(\curlyR)$ is empty 
and so \lb  
\,$\Thms(\curlyL,\varphi,\curlyR) = \Thms(\Fact(\curlyR))$. 
Also if \,$S \!\subseteq\! \Thms(\Fact(\curlyR))$\, 
then \,$\Fact(\curlyR) \!\cup\! S$\, is satisfiable. 
So (the proof algorithm of) classical logic is strongly $\aleph_0$-consistent. 

Contradictions are not plausible. 
Therefore plausible proof algorithms must be 1-consistent. 
If $s \e \Thms(\curlyL,\alpha,\curlyR)$\, 
then in the situation defined by $\curlyR$, $s$ is more likely to be true than not. 
Therefore \,$\Fact(\curlyR)\!\cup\!\{s\}$\, is satisfiable. 
Hence, plausible proof algorithms must be strongly 1-consistent. 

Now consider strong 2-consistency. 
So suppose $f$ and $g$ are sentences such that 
\,$\{f,g\} \!\subseteq\! \Thms(\curlyL,\alpha,\curlyR)$. 
By strong 1-consistency, we have that both 
\,$\Fact(\curlyR)\!\cup\!\{f\}$\, and \,$\Fact(\curlyR)\!\cup\!\{g\}$\, are satisfiable. 
So $f$ is not a contradiction and $g$ is not a contradiction. 
If \,$\Fact(\curlyR) \!\cup\! \{f,g\}$\, is unsatisfiable then 
$f$ is not a tautology and $g$ is not a tautology. 
Also \,$\Fact(\curlyR) \!\cup\! \{f\} \imps \neg g$. 
Hence \,$\neg g \in \Thms(\Fact(\curlyR\!+\!f))$. 
There are now two cases. 

If \,$\neg g = \AND\{s,\bbt\}$\, where 
\,$s \e \Thms(\curlyL,\varphi,\curlyR\!+\!f)$\, and \,$\bbt \e \Taut$\, 
then \,$s \equiv \neg g$.  
By Principle~\ref{Prin:Plausible Right Weakening} (Plausible Right Weakening), 
\,$s \e \Thms(\curlyL,\alpha,\curlyR)$. 
Otherwise, by Principle~\ref{Prin:Factual Subclassicality} (Factual Subclassicality), 
\,$\neg g \e \Thms(\curlyL,\varphi,\curlyR\!+\!f)$, and so by  
Principle~\ref{Prin:Plausible Right Weakening} (Plausible Right Weakening), 
\,$\neg g \e \Thms(\curlyL,\alpha,\curlyR)$. 

So in both cases we have 
\,$\{g,s\} \!\subseteq\! \Thms(\curlyL,\alpha,\curlyR)$\, where \,$s \equiv \neg g$. 
Although `likely' is an imprecise concept, it has the property that for any formula $g$, 
if \,$s \equiv \neg g$\, then at most one of $g$ and $s$ is likely. 
Therefore we do not have \,$\{g,s\} \!\subseteq\! \Thms(\curlyL,\alpha,\curlyR)$. 
This contradiction shows that \,$\Fact(\curlyR) \!\cup\! \{f,g\}$\, is satisfiable. 
So plausible proof algorithms are strongly 2-consistent. 
Hence our first consistency principle is necessary. 

%Principle~3.25
\begin{Prin}[The Strong 2-Consistency Principle] 
\label{Prin:Strong 2-Consistency} 
%\raggedright \parindent = 1.2em
A plausible proof algorithm must be strongly 2-consistent. 
\end{Prin}

Consider the 3-lottery example (Example~\ref{Eg:3-lottery}). 
Let \,$U = \{\neg s_1, \neg s_2, \OR\{s_1, s_2\}\}$. 
Then for each $x$ in $U$, $x$ is likely; and $\neg x$ is not likely. 
But $U$ is unsatisfiable. 
Hence the necessity of our second consistency principle. 

%Principle~3.26
\begin{Prin}[The Non-3-Consistency Principle] 
\label{Prin:Non-3-Consistency} 
%\raggedright \parindent = 1.2em
A plausible proof algorithm that can prove disjunctions must not be 3-consistent. 
\end{Prin}
%%%%%%%%%%%%%%%%%%%%%%%%%%%%%%%%%%%%%%%%%

%Subsection 3.9
\subsection{Different Intuitions: Ambiguity} 
\label{PLPR:Ambiguity}
With the possible exception of sentences involving tautologies, 
classical logic captures our intuition about what follows from a satisfiable set of facts. 
But there are different well-informed intuitions about 
what follows from a plausible-reasoning situation, 
depending on whether ambiguity is blocked or propagated. 
This is shown by our fourth signpost example. 

%Example~3.27
\begin{Eg}[The Ambiguity Puzzle] \ 
\label{Eg:Ambiguity}
%\raggedright \parindent = 1.2em
\begin{compactenum}[1)]
\item $a$ is a fact. \ $b$ is a fact. \ $c$ is a fact. 
\item If $a$ is true then $e$ is likely. 
\item If $d$ is true then $\neg e$ is likely. 
\item If $b$ is true then $d$ is likely. 
\item If $c$ is true then $\neg d$ is likely. 
\end{compactenum}
\end{Eg}

The Ambiguity Puzzle can be represented by the following diagram. 

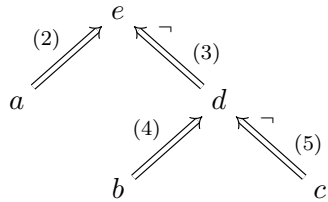
\vspace{2ex}
\begin{center}
%\ar = \arrow. tikzcd works on a grid system like tabular. 
\begin{tikzcd}[cramped] 
  & e \ar[Leftarrow, rd, "\neg" near start, "(3)" near end] &  &  \\
a \ar[Rightarrow, ru, "(2)"] &  & d \ar[Leftarrow, rd, "\neg" near start, "(5)" near end] & \\
  & b \ar[Rightarrow, ru, "(4)"] &  & c
\end{tikzcd}
\end{center}
\vspace{2ex}

What can be concluded about $e$? 
The evidence for $e$ is $a$ and (2). 
The evidence against $e$ comes from $d$ and (3), and $b$ and (4). 
If we knew that $d$ was definitely true then 
the evidence for $e$ and against $e$ would be equal. 
Ignoring (5), $d$ is only likely by (4), 
so the evidence against $e$ is weaker than the evidence for $e$. 
But $c$ and (5) means that $d$ is even less likely, 
and so the evidence against $e$ has been further weakened. 
Thus $e$ is more likely than $\neg e$. 

So reasoning that is based on the `best bet' or the `most likely' 
or the `balance of probabilities' concludes $e$. 
However, $d$ may be true, and so it is reasonable 
to have some doubts about the truth of $e$. 
So reasoning that is `beyond reasonable doubt' does not conclude $e$. 

A sentence $s$ is said to be \defn{ambiguous} iff 
there is evidence for $s$ and 
there is evidence against $s$ and 
neither $s$ nor $\neg s$ can be proved.
Since $b$ and (4) and $c$ and (5) give equal evidence for and against $d$, 
$d$ is ambiguous. 

If the evidence against $e$ has been weakened sufficiently 
to allow $e$ to be concluded, then $e$ is not ambiguous. 
So the ambiguity of $d$ has been blocked from propagating to $e$. 
An algorithm that can prove $e$ is said to be \defn{ambiguity blocking}. 
This level of reasoning is appropriate if 
the benefit of being right outweighs the penalty for being wrong. 

If the evidence against $e$ has not been weakened sufficiently 
to allow $e$ to be concluded, then $e$ is ambiguous. 
So the ambiguity of $d$ has been propagated to $e$. 
An algorithm that cannot prove $e$ is said to be \defn{ambiguity propagating}. 
This more cautious level of reasoning is appropriate if 
the penalty for being wrong outweighs the benefit of being right. 

The Anglo-American legal system uses a hierarchy of proof levels. 
(For example see page 584 of \cite{Mann2013}.)
Two of these levels are the `balance of probabilities' (used in civil cases) 
which, as noted above, is ambiguity blocking; 
and `beyond reasonable doubt' (used in criminal cases) 
which, as noted above, is ambiguity propagating. 
So there is a need for a proof algorithm that blocks ambiguity and 
one that propagates ambiguity. 

To avoid confusion, one should know which algorithm is used; 
unless it is irrelevant to the point being made. 
This, and our observation at the end of 
Subsection~\ref{PLPR:Representation}~Representation, that 
a logic for plausible reasoning should have a factual proof algorithm, leads to our next principle. 
A logic for plausible reasoning that does not have both an ambiguity blocking 
algorithm and an ambiguity propagating will get the answer to the 
Ambiguity Puzzle (Example~\ref{Eg:Ambiguity}) wrong for some cases. 
So the next principle is regarded as necessary. 

%Principle~3.28
\begin{Prin}[The Many Proof Algorithms Principle] 
\label{Prin:Many Proof Algorithms} 
%\raggedright \parindent = 1.2em
A logic for plausible reasoning should have at least 
\begin{compactenum}[1)]
\item a factual proof algorithm, 
\item an ambiguity blocking plausible proof algorithm, and 
\item an ambiguity propagating plausible proof algorithm. 
\end{compactenum}
Also, the proof algorithm used to prove a sentence must be explicit. 
\end{Prin}

Clearly the algorithms in (2) and (3) must be different. 
%%%%%%%%%%%%%%%%%%%%%%%%%%%%%%%%%%%%%%%%%

%Subsection 3.10
\subsection{Termination} 
\label{PLPR:Termination}
Proof algorithms either terminate after a finite number of steps or they do not terminate. 
Proof algorithms that do not terminate are usually regarded as useless. 
But this need not be the case. 
It can happen that, although the proof algorithm does not terminate, 
after a finite number of steps it is clear that what is to be proved is in fact unprovable. 
This realisation leads to the following definition. 

%Definition~3.29
\begin{Defn} \label{Defn:decisive} 
%\raggedright \parindent = 1.2em
Suppose $\alpha$ is a proof algorithm and $s$ is a sentence. 
A proof of $s$ using $\alpha$ is called an \defn{$\alpha$-proof} of $s$. 
\begin{compactenum}[1)] 
\item %1) 
	$\alpha$ is \defn{decisive for} $s$ iff after a finite number of steps \nl 
	either it is clear that there is an $\alpha$-proof of $s$, \nl 
	or it is clear that there is not any $\alpha$-proof of $s$. 
\item %2) 
	$\alpha$ is \defn{decisive} iff for all sentences $s$, $\alpha$ is decisive for $s$. 
\end{compactenum} 
\end{Defn} 

It may happen that $\alpha$ terminates in a state indicating that it made a choice that 
did not lead to an $\alpha$-proof of $s$, 
but there may be a choice that does lead to an $\alpha$-proof of $s$. 
So $\alpha$ is not decisive for $s$. 
But it seems a bit harsh to declare that a proof algorithm is not a plausible proof algorithm 
just because it is not decisive. 
Hence our next principle is clearly desirable, but not necessary. 

%Principle~3.30
\begin{Prin}[The Decisiveness Principle] 
\label{Prin:Decisiveness} 
%\raggedright \parindent = 1.2em
Factual and plausible proof algorithms should be decisive. 
\end{Prin}

This Decisiveness Principle is a weakening of the Decisiveness Principle on page 12 of 
\cite{Billington2017}, page 138 of \cite{Billington2019}, and in \cite{Billington2024}. 
However, the old Decisiveness Principle implied that loops were undesirable, 
but the new one does not. 

A proof algorithm that gets into a loop will not terminate. 
It is highly desirable to prevent such loops, particularly if the algorithm is to be implemented. 
So our next principle is desirable. 

%Principle~3.31
\begin{Prin}[The No Loops Principle] 
\label{Prin:No Loops} 
%\raggedright \parindent = 1.2em
\index{03.10.03@\ref{Prin:Decisiveness} Principle}
\index{No Loops Principle~\ref{Prin:No Loops}} 
Factual and plausible proof algorithms should not get into loops. 
\end{Prin}
%%%%%%%%%%%%%%%%%%%%%%%%%%%%%%%%%%%%%%%%%

%Subsection 3.11
\subsection{Truth Values} 
\label{PLPR:Truth Values}
Let us change our focus from deduction to the more semantic notion 
of assigning truth values to statements. 
For classical propositional logic there are exactly two truth values: 
$\T$ for true and $\F$ for false. 
If $v$ is a valuation and $f$ and $g$ are formulas then 
\begin{compactenum}[1)] 
\item \label{Excluded Middle} %1) 
	Either \,$v(f) = \T$\, or \,$v(\neg f) = \T$\, but not both, 
	(the Excluded Middle property) and 
\item \label{AND is true} %2) 
	$v(\AND\{f,g\}) = \T$ \ iff \ $v(f) = \T = v(g)$, and 
\item \label{OR is true} %3) 
	$v(\OR\{f,g\}) = \T$ \ iff \ either \,$v(f) = \T$\, or \,$v(g) = \T$\, (or both). 
\end{compactenum}

The 3-lottery example (Example~\ref{Eg:3-lottery}) shows that 
the closest plausible reasoning can get to (\ref{AND is true}) and 
(\ref{OR is true}) is (\ref{half AND}) and (\ref{half OR}) below. 
\begin{compactenum}[1)] 
\addtocounter{enumi}{3}
\item \label{half AND} %4) 
	If \,$v(\AND\{f,g\}) = \T$\, then \,$v(f) = \T = v(g)$. 
\item \label{half OR} %5) 
	If \,$v(f) = \T$\, or \,$v(g) = \T$\, then \,$v(\OR\{f,g\}) = \T$. 
\end{compactenum}

Intuitively, some formulas concerning the 3-lottery example have different truth values; 
for example $\OR\{s_1, s_2, s_3\}$ is definitely true, 
$\neg \OR\{s_1, s_2, s_3\}$ is definitely false, $\neg s_1$ is probably true, and 
$s_1$ is probably false. 
But consider our fifth signpost example. 

%Example~3.32
\begin{Eg}[The 2-lottery example] 
\label{Eg:2-lottery}
%\raggedright \parindent = 1.2em
Consider the 2-lottery based on $\{1,2\}$. 
Then we have the following two statements. 
\begin{compactenum}[1)] %\raggedright
\item Exactly one element of $\{s_1, s_2\}$ is true. 
\item Each element of $\{s_1, s_2\}$ is not probably true and not probably false. 
\end{compactenum}
\end{Eg}

Intuitively, the formula $s_1$ of Example~\ref{Eg:2-lottery} is as likely to be true as false. 
So plausible reasoning distinguishes between at least 3 truth values: \nl
one indicating that a formula is more likely to be true than false, \nl
one indicating that a formula is as likely to be true as false, and \nl
one indicating that a formula is more likely to be false than true. \nl 
Two more truth values could be added. 
One indicating that a formula is certainly true, and one indicating that a formula is certainly false. 

Therefore a logic with only two truth values cannot fully express all 
the different possibilities that exist in plausible reasoning. 
Hence the following necessary principle of plausible reasoning. 

%Principle~3.33
\begin{Prin}[The Included Middle Principle] 
\label{Prin:Included Middle} 
%\raggedright \parindent = 1.2em
If a logic for plausible reasoning has a truth theory 
then that truth theory must have at least 3 truth values. 
\end{Prin}

The Included Middle Principle (Principle~\ref{Prin:Included Middle}) 
is our last formal principle of a logic for plausible reasoning. 
However, there is another informal, but highly desirable, 
principle of a logic for plausible reasoning that we shall consider in the next section. 
%%%%%%%%%%%%%%%%%%%%%%%%%%%%%%%%%%%%%%%%%

%Subsection 3.12
\subsection{Correctness} 
\label{PLPR:Correctness}
A logic that satisfies all the previous principles could nonetheless have a fatal flaw. 
It could give an unsatisfactory answer to a particular example. 
Some examples may well have no set of answers that are generally agreed upon. 
But some examples, in particular the signpost examples, 
do have a set of answers that are generally agreed upon. 
We might call these answers the correct answers. 
So it is tempting to state a principle of correctness similar to the following statement. 
``When correct answers exist, a logic for plausible reasoning must give 
all the correct answers, and no incorrect answers."

The problem with such a principle is that it is impossible to show that any logic satisfies it. 
The most that can be done is to produce a counter-example that shows a logic fails the principle, 
or demonstrate that for a chosen set of examples the logic gets the correct answers. 
But there might exist an example that shows the logic fails the principle of correctness. 

Thus we shall refrain from trying to formally state a Correctness Principle, 
even though such a principle is highly desirable. 

%2345678901234567890123456789012345678901234567890123456789012345678901234567890
%Section 4
\section{Plausible Logic (PL)} 
\label{Section:PL}
PL deduces conclusions about plausible-reasoning situations that 
may contain information that is certain, or factual, as well as information that is plausible. 
Definitions and membership of categories are facts; for example `Whales are mammals'. 
The plausible information is represented by defeasible rules and warning rules. 

Intuitively the various kinds of rules have the following meanings. 
The strict rule \,$A \!\strArr\! c$\, means 
if every formula in $A$ is accepted then $c$ is acceptable. 
So strict rules are like material implication except that 
$A$ is a finite set of formulas rather than a single formula. 
(We have already seen that $A$ and $\AND A$ behave differently.)
For example, `nautiluses are cephalopods' could be written as 
\,$\{n(v)\} \!\strArr\! c(v)$. 

The defeasible rule \,$A \!\defArr\! c$\, means 
if every formula in $A$ is accepted and there is no evidence against $c$, then $c$ is likely. 
For example, `molluscs usually have external shells' could be written as 
\,$\{m(v)\} \!\defArr\! s(v) $, and `cephalopods usually have no external shells' 
could be written as \,$\{c(v)\} \!\defArr\! \neg s(v)$. 

The warning rule \,$A \!\warnArr\! c$\, roughly means that if every formula in $A$ is accepted 
and there is no evidence against $c$ then $c$ might be acceptable. 
So \,$A \!\warnArr\! \neg c$\, warns against concluding usually $c$, 
but does not support usually $\neg c$. 
Warning rules can be used to prevent conclusions that are too risky. 
For example, `scared quails might fly' could be written as \,$\{q(v), s(v)\} \warnArr f(v)$. 
The idea is that being scared is not enough evidence to conclude that a quail will probably fly, 
but it is enough evidence to prevent concluding that it probably will not fly. 
Warning rules can be used to prevent unwanted chaining. 
For example, suppose we have `if $a$ then usually $b$' ($\{a\} \!\defArr\! b$) 
and `if $b$ then usually $c$' ($\{b\} \!\defArr\! c$). 
Then it may be too risky to conclude `usually $c$' from $a$. 
Without introducing evidence for $\neg c$, 
the conclusion of `usually $c$' from $a$ can be prevented by 
the warning rule \,$\{a\} \!\warnArr\! \neg c$. 
An instance of this example can be created by letting 
$a$ be \,$v \e \{1,2,3,4\}$, $b$ be \,$v \e \{2,3,4\}$, and $c$ be \,$v \e \{3,4,5\}$. 
Warning rules have also been called `defeaters' and `interfering rules'. 
%%%%%%%%%%%%%%%%%%%%%%%%%%%%%%%%%%%%%%%%%

%Subsection 4.1
\subsection{Plausible-Descriptions} 
\label{PL:Plausible-Descriptions}
Let us start the formalities by defining an alphabet for PL. 
%Definition~4.1
\begin{Defn} \label{Defn:alphabetPL} 
%\raggedright \parindent = 1.2em
An \defn{alphabet for PL} consists of the following 7 pairwise disjoint sets. 
\begin{compactenum}[1)]
\item \label{PredPL} %1)
	A non-empty countable set, $\Pred$, of \defn{predicate symbols}. 
	If \,$n \e \ZZ^+$\, then an $n$-ary predicate symbol represents 
	an $n$-ary relation. 
\item \label{FuncPL} %2)
	A countable set, $\Func$, of \defn{function symbols}. 
	If \,$n \e \ZZ^+$\, then an $n$-ary function symbol represents an $n$-ary function. 
\item \label{ConstPL} %3)
	A countable set, $\Const$, of \defn{constant symbols}. 
\item \label{VarPL} %4)
	A set, $\Var$, of \defn{variable symbols} such that 
	either $\Var$ is countably infinite, or $\Var$ is empty. 
	We usually abbreviate ``variable symbol" to ``variable". 
\item \label{quantifiersPL} %5)
	The set, \,$\{\forall, \exists\}$, of \defn{quantifiers}. 
	$\forall$ is called the \defn{universal quantifier}, and 
	$\exists$ is called the \defn{existential quantifier}.
\item \label{connectivesPL} %6)
	The set $\{\neg, \AND, \OR, \strArr, \defArr, \warnArr\}$ of 
	\defn{connectives}  denoting negation, conjunction, disjunction, 
	the strict arrow, the defeasible arrow, and the warning arrow respectively. 
\item \label{priority relation symbol} %7)
	The set $\{>\}$. The symbol $>$ denotes a priority relation. 
\item \label{punctuation symbolsPL} %8)
	The set of punctuation symbols consisting of the comma, 
	both parentheses (round brackets), and both braces (curly brackets). 
\end{compactenum}
\end{Defn} 

Rules, and their associated terms, are defined next. 

%Definition~4.2
\begin{Defn} \label{Defn:rule}
%\raggedright \parindent = 1.2em
A \defn{rule}, $r$, consists of the following three parts. 
\begin{compactitem}[\textbullet]
\item A finite, possibly empty, set $A(r)$ of formulas, called the \defn{set of antecedents} of $r$. 
\item One of the following three arrows: $\strArr, \,\defArr$, and $\warnArr$. 
\item A formula $c(r)$ called the \defn{consequent} of $r$. 
\end{compactitem}
There are three kinds of rules as defined below. 
\begin{compactenum}[i)]
\item A \defn{strict rule} $r$ uses the \defn{strict arrow}, $\strArr$, and is written 
	\,$A(r) \strArr c(r)$. 
\item A \defn{defeasible rule} $r$ uses the \defn{defeasible arrow}, $\defArr$, and is written 
	\,$A(r) \defArr c(r)$. 
\item A \defn{warning rule} $r$ uses the \defn{warning arrow}, $\warnArr$, and is written 
	\,$A(r) \warnArr c(r)$. 
\end{compactenum}
If $r$ is any rule then define \,$\Var(r) = \{v \e \Var : v$ occurs in $r\}$. 
\end{Defn} 

So $\Var(r)$ is the set of variables that occur in $r$. 
The following sets of rules are useful. 

%Definition~4.3
\begin{Defn} \label{Defn:sets of rules}
%\raggedright \parindent = 1.2em
Let $R$ be any set of rules. 
\begin{compactenum}[1)]
\item $R_s$ is the set of strict rules in $R$. That is, \,$R_s = \{r \e R : r$ is a strict rule$\}$. 
\item $R_d$ is the set of defeasible rules in $R$. That is, 
	\,$R_d = \{r \e R : r$ is a defeasible rule$\}$. 
\item $R_w$ is the set of warning rules in $R$. That is, 
	\,$R_w = \{r \e R : r$ is a warning rule$\}$. 
\item $R_{sd} = R_s \!\cup\! R_d$.
\end{compactenum}
\end{Defn} 

When a substitution $\sigma$ is applied to a rule $r$, 
we want the variables not in $r$ to remain unchanged by $\sigma$. 
Define $\Doc(\sigma)$, the \defn{domain of change} of $\sigma$, by 
\,$\Doc(\sigma) = \{v \e \Var : v\sigma \neq v\}$\, to be the set of variables that 
$\sigma$ changes. 
The set of all substitutions is denoted by $\Sigma$. 

%Definition~4.4
\begin{Defn} \label{Defn:RSigma etc}
%\raggedright \parindent = 1.2em
Let $R$ be any set of rules. 
If $r$ is a rule and $\sigma$ is a substitution then we say $r\sigma$ is a \defn{rule-instance}. 
If \,$\Var = \{\}$\, then define \,$R\Sigma = R$. 
If \,$\Var \neq \{\}$\, then define \,$R\Sigma = 
\{r\sigma : r \e R, \sigma \e \Sigma$, and \,$\Doc(\sigma) \!\subseteq\! \Var(r)\}$. 
\end{Defn} 

A priority relation, $>$, on $R\Sigma$ is used to indicate the more relevant of two rule-instances. 
For example, when reasoning about the external appearance of cephalopods, 
any instance of the specific rule `cephalopods usually do not have external shells', 
\,$\{c(v)\} \!\defArr\! \neg s(v)$, 
is more relevant than the corresponding instance of the general rule 
`molluscs usually have external shells', \,$\{m(v)\} \!\defArr\! s(v)$. 
So if $b$ is a cephalopod then 
\,$\{c(b)\} \!\defArr\! \neg s(b) \ > \ \{m(b)\} \!\defArr\! s(b)$. 
Some common policies for defining $>$ are: 
prefer specific rule-instances over general rule-instances; 
prefer authoritative rule-instances, (for example national laws override state laws); 
prefer recent rule-instances (as they are more up-to-date); and 
prefer more reliable rule-instances. 
If \,$\{r\rho, s\sigma\} \subseteq R\Sigma$\, and \,$r\rho > s\sigma$\, then  
$r\rho$ is \defn{superior} to $s\sigma$ and $s\sigma$ is \defn{inferior} to $r\rho$. 

In Section~\ref{Section:Principles}, Principles of Logics for Plausible Reasoning, 
a plausible-reasoning situation was specified by a plausible-representation 
\,$\curlyR = (\Fact(\curlyR), \Plaus(\curlyR))$, where $\Fact(\curlyR)$ was a satisfiable set of 
sentences representing the factual part of $\curlyR$, 
and $\Plaus(\curlyR)$ was a set representing the plausible part of $\curlyR$. 
Plausible Logic reasons with a special kind of plausible-representation, 
called a plausible-description. 
A plausible-description, $\curlyD = (\Ax,R,>)$, has the following three parts: 
a satisfiable set of sentences, $\Ax$, called the axioms of $\curlyD$; 
a set of rules $R$; and a priority relation $>$ on $R\Sigma$. 
However, each of these parts must satisfy some extra conditions, which are specified below. 

By Skolemisation, we may suppose that each sentence in $\Ax$ is a \defn{basic sentence}; 
that is, the universal closure of a clause, written $\forall c$. 
So if \,$a \e \Ax$\, then there is a clause $c$ such that \,$a = \forall c$. 
If $c$ is a clause then define \,$\Cl(\forall c) = c$\, and \,$\Cl(\Ax) = \{\Cl(a) : a \e \Ax\}$. 

It is useful to convert an axiom with $n$ literals into \,$2^n - 1$\, strict rules. 
The conversion is done by the function $\Rul(.)$ in the usual way, 
as shown by the following example. \nl
$\Rul(\OR\{l_1,l_2,l_3\}) = \{ \quad \{\} \strArr \OR\{l_1,l_2,l_3\},$ \nl 
\begin{tabular}{@{}p{9em}p{9em}l}
$\{\nnot l_1\} \strArr \OR\{l_2,l_3\}$,
	& $\{\nnot l_2\} \strArr \OR\{l_1,l_3\}$,
	& $\{\nnot l_3\} \strArr \OR\{l_1,l_2\}$,\\
$\{\AND\{\nnot l_2,\nnot l_3\}\} \strArr l_1$,
	& $\{\AND\{\nnot l_1,\nnot l_3\}\} \strArr l_2$,
	& $\{\AND\{\nnot l_1,\nnot l_2\}\} \strArr l_3$ \quad $\}$. 
\end{tabular} \nl
The full definition of $\Rul(.)$ is given in the next definition. 

%Definition~4.5
\begin{Defn} \label{Defn:Rul(.)} 
%\raggedright \parindent = 1.2em
Let $L$ be any non-empty finite set of literals. 
Suppose $l$ is a literal. 
There are several cases depending on the cardinality of $L$. 
\begin{compactenum}[1)]
\item %1)
	$\Rul(\OR\{l\}) = \Rul(l) = \{ \ \{\} \strArr l\}$. 
\item %2) 
	If \,$|L| = 2$\, then \,$\Rul(\OR L) = 
	\{ \ \{\} \strArr \OR L\} \ \cup \ 
	\{ \nnot K \strArr l : L = K\!\cup\!\{l\}$\, and \,$l \nte K\}$. 
\item %3)
	If \,$|L| = 3$\, then \,$\Rul(\OR L) = 
	\{ \ \{\} \strArr \OR L\} \ \cup \nl 
	\{ \ \{\nnot l\} \strArr \OR K : L = K\!\cup\!\{l\}$\, and 
	                                 \,$l \nte K \} \ \cup 
	\{ \ \{\AND\nnot(L\!-\!\{l\})\} \strArr l : l \e L\}$. 
\item %4)
	If \,$|L| \geq 4$\, then \,$\Rul(\OR L) = 
	\{ \ \{\} \strArr \OR L\} \ \cup \nl 
	\{ \ \{\nnot l\} \strArr \OR K : L = K\!\cup\!\{l\}$\, and 
	                                 \,$l \nte K \} \ \cup \nl 
	\{ \ \{\AND\nnot(L\!-\!K)\} \strArr \OR K : K\!\subset\!L$\, and 
		\,$|L\!-\!K| \geq 2$\, and \,$|K| \geq 2\} \ \cup \nl
	\{ \ \{\AND\nnot(L\!-\!\{l\})\} \strArr l : l \e L\}$. 
\item %5)
	If $C$ is a set of clauses then \,$\Rul(C) = \bigcup\{\Rul(c) : c \e C\}$. 
\end{compactenum}
\end{Defn} 

To prevent a formula, $f$, and its negation $\neg f$, both being proved to be plausible, 
$>$ must satisfy a finiteness condition 
(see Definition~\ref{Defn:priority relation}(\ref{R[f] = ...},\ref{well-founded on}) below). 

%Definition~4.6
\begin{Defn} \label{Defn:priority relation} 
%\raggedright \parindent = 1.2em
Let $R$ be any set of rules, and $f$ be any formula. 
\begin{compactenum}[1)]
\item \label{R[f] = ...} %1) 
	$R[f] = \{r\rho \in R\Sigma : \Ax \cup \{c(r\rho)\} \imps f, 
	\ \Ax \cup \{c(r\rho)\}$ is satisfiable, and \,$\Ax \nimps f \}$. 
\item \label{well-founded on} %2) 
	$>$ is \defn{well-founded on} $f$ iff there does not exist an infinite sequence \nl 
	$(r_1\rho_1, s_2\sigma_2, r_3\rho_3, s_4\sigma_4, ...)$ such that for each $i$, 
	\,$r_i\rho_i \e R_{sd}[f]$\, and \,$s_i\sigma_i \e R_{sd}[\neg f]$, and for each odd $i$, 
	\,$r_{i+2}\rho_{i+2} > s_{i+1}\sigma_{i+1} > r_{i}\rho_{i}$. 
\item \label{priority relation} %3) 
	$>$ is a \defn{priority relation} on $R\Sigma$ iff 
	\,$> \ \subseteq R\Sigma \!\times\! R\Sigma$\, and $>$ is well-founded on every formula. 
\end{compactenum}
\end{Defn} 

The infinite sequence in Definition~\ref{Defn:priority relation}(\ref{well-founded on}) 
may have repeated elements. 
If $>$ is empty then $>$ is a priority relation on $R\Sigma$. 
A priority relation on $R\Sigma$ does not have to be transitive; indeed it is usually atransitive. 

We can now formally define a plausible-description. 

%Definition~4.7
\begin{Defn} \label{Defn:plausible-description} 
%\raggedright \parindent = 1.2em
$\curlyD$ is a \defn{plausible-description} iff the following 5 conditions all hold. 
\begin{compactenum}[1)]
\item %1)
	$\curlyD = (\Ax,R,>)$. 
\item %2) 
	$\Ax$ is a satisfiable set of basic sentences. 
	We say $s$ is an \defn{axiom of} $\curlyD$ iff \,$s \e \Ax$. 
\item %3)
	$R$ is a set of rules. 
\item %4)
	$R_s = \Rul(\Cl(\Ax))$. 
\item %5) 
	$>$ is a \defn{priority relation} on $R\Sigma$. 	
\end{compactenum}
\end{Defn} 
%%%%%%%%%%%%%%%%%%%%%%%%%%%%%%%%%%%%%%%%%

%Subsection 4.2
\subsection{The Proof Theory} 
\label{PL:Proof Theory}
In this subsection we define the proof function and the proof algorithms of Plausible Logic (PL) 
given a plausible-description. 
This complex task will be accomplished by stating the top level plan, 
and then progressively refining this plan. 

The Representation Principle, Principle~\ref{Prin:Representation}, 
says that PL must distinguish between facts and non-facts. 
So our top level plan for proving a formula is the following. 

\begin{flushleft} 
\textbf{Plan} Distinguish between proving formulas that are facts and 
proving formulas that are not facts. 
\end{flushleft}

Let \,$(\Ax,R,>)$\, be any plausible-description. 
We shall regard any formula, $f$, such that \,$\Ax \imps f$\, as a fact.  

Lower case Greek letters will be used to denote the proof algorithms that 
we shall eventually define. 
A general proof algorithm will be denoted by $\alpha$ (a for alpha and algorithm). 
We shall use $\varphi$ (f for phi and fact) to denote our factual proof algorithm. 
Until a further refinement is needed we shall use the notation \,$\alpha \proves f$\, 
to denote that a formula $f$ is proved by the proof algorithm $\alpha$. 

In the situation defined by \,$(\Ax,R,>)$, facts are always true, 
and so we shall say that they are (at least) probably true. 
Hence we shall decree that facts are provable by all proof algorithms. 
That is, if \,$\Ax \imps f$\, then \,$\alpha \proves f$. 
The factual algorithm proves a formula iff it is a fact. 
In symbols, \,$\varphi \proves f$\, iff \,$\Ax \imps f$. 

Now consider formulas $f$ that are not facts; that is, \,$\Ax \nimps f$. 
To (plausibly) prove $f$ we need to do two things. 
First, establish some evidence for $f$. 
Second, defeat all the evidence against $f$. 
This will satisfy the requirements of the Evidence Principle, Principle~\ref{Prin:Evidence}. 
So our first refinement of the plan is the following. 

%Refinement~4.8 
\begin{Refine} \label{Refine:ProofRelationFirst} 
%\raggedright \parindent = 1.2em
Suppose \,$(\Ax,R,>)$\, is a plausible-description, $\alpha$ is a proof algorithm, and 
$f$ is a formula. 
\begin{compactenum}[1)]
\item \label{refine1:fact} %1)
	If \,$\Ax \imps f$\, then \,$\alpha \proves f$. \ 
	Also \,$\varphi \proves f$\, iff \,$\Ax \imps f$. 
\item \label{refine1:plaus} %2)
	If \,$\Ax \nimps f$\, and \,$\alpha \neq \varphi$\, then \,$\alpha \proves f$\, iff both 
	(\ref{refine1:plaus}.\ref{refine1:For}) and (\ref{refine1:plaus}.\ref{refine1:Against}) hold. 
	\begin{compactenum}[\ref{refine1:plaus}.1)]
	\item \label{refine1:For} %2.1)
		Some evidence for $f$ is established. 
	\item \label{refine1:Against} %2.2)
		All the evidence against $f$ is defeated. 
	\end{compactenum}
\end{compactenum}
\end{Refine} 

In Refinement~\ref{Refine:ProofRelationFirst}(\ref{refine1:plaus}.\ref{refine1:For}) 
the evidence for $f$ consists of strict or defeasible rules that have a consequent that implies $f$. 
Therefore if $r$ is a strict or defeasible rule and \,$c(r) \imps f$\, then $r$ is evidence for $f$. 
This evidence for $f$ is `established' by proving $A(r)$, the set of antecedents of $r$. 
But $A(r)$ is a finite set of formulas, not a formula. 
By proving a finite set $F$ of formulas we shall mean proving every formula in $F$. 
In symbols, \,$\alpha \proves F$ \ iff \ for all $f$ in $F$, \,$\alpha \proves f$. 
So if $F$ is empty we have \,$\alpha \proves \{\}$. 

Since the axioms are always true, we can weaken the requirement that \,$c(r) \imps f$\, to 
\,$\Ax \cup \{c(r)\} \imps f$\, provided we are careful. 
If \,$\Ax \cup \{c(r)\}$\, is unsatisfiable then we do not want to regard $r$ as supporting $f$. 

The evidence for a formula $f$ is defined next. 

%Definition~4.9
\begin{Defn} \label{Defn:R'[f]} 
%\raggedright \parindent = 1.2em
Suppose $(\Ax,R,>)$ is a plausible-description, \,$R' \!\subseteq\! R$, 
\,$s\sigma \e R\Sigma$, and $f$ is a formula. 
(We repeat Definition~\ref{Defn:priority relation}(\ref{R[f] = ...}).) \nl 
$R'[f] = \{r\rho \in R'\Sigma : \Ax \cup \{c(r\rho)\} \imps f, 
\ \Ax \cup \{c(r\rho)\}$ is satisfiable, and \,$\Ax \nimps f \}$. \nl
\,$R'[f]_{>s\sigma} = \{t\tau \e R'[f] : t\tau > s\sigma\}$. \nl
$R_{sd}[f]$ is the set of evidence for $f$. 
$R[\neg f]$ is the set of evidence against $f$. 
\end{Defn} 

Collecting these ideas together gives our second refinement. 

%Refinement~4.10 
\begin{Refine} \label{Refine:ProofRelationSecond} 
%\raggedright \parindent = 1.2em
Suppose \,$(\Ax,R,>)$\, is a plausible-description, $\alpha$ is a proof algorithm, and 
$f$ is a formula. 
\begin{compactenum}[1)]
\item \label{refine2:set} %1)
If $F$ is a finite set of formulas then 
\,$\alpha \proves F$ \ iff \ for all $f$ in $F$, \,$\alpha \proves f$. 

\item \label{refine2:fact} %2)
If \,$\Ax \imps f$\, then \,$\alpha \proves f$. \ 
Also \,$\varphi \proves f$\, iff \,$\Ax \imps f$. 

\item \label{refine2:fml} %3)
If \,$\Ax \nimps f$\, and \,$\alpha \neq \varphi$\, 
then \,$\alpha \proves f$\, iff there exists $r\rho$ in $R_{sd}[f]$ 
such that both (\ref{refine2:fml}.\ref{refine2:For}) and 
(\ref{refine2:fml}.\ref{refine2:Against}) hold. 
	\begin{compactenum}[\ref{refine2:fml}.1)]
	\item \label{refine2:For} %3.1)
	$\alpha \proves A(r\rho)$. 

	\item \label{refine2:Against} %3.2)
	All the evidence against $f$ is defeated. 
	\end{compactenum}

\end{compactenum}
\end{Refine} %4.10

Suppose \,$s\sigma \e R[\neg f]$. 
Then $\alpha$ defeats $s\sigma$ by either team defeat or by disabling $s\sigma$. 
The team of rule-instances for $f$ is $R_{sd}[f]$. 
The rule-instance $s\sigma$ is defeated by $\alpha$ using \defn{team defeat} iff 
there is a rule-instance $t\tau$ in $R_{sd}[f]$ such that 
$t\tau$ is superior to $s\sigma$, \,$t\tau > s\sigma$, and $\alpha$ proves 
the set of antecedents of $t\tau$, \,$\alpha \!\proves\! A(t\tau)$. 
Also $s\sigma$ is \defn{disabled} by $\alpha$ iff $\alpha$ can prove that 
the set of antecedents of $s\sigma$, $A(s\sigma)$, is not provable by $\alpha$. 
This is a stronger condition than \,$\alpha \nproves A(s\sigma)$. 
Collecting these ideas together gives our third refinement. 

%Refinement~4.11 
\begin{Refine} \label{Refine:ProofRelationThird} 
%\raggedright \parindent = 1.2em
Let \,$(\Ax,R,>)$\, be a plausible-description, $\alpha$ be a proof algorithm, and $f$ be a formula. 
\begin{compactenum}[1)]
\item \label{refine3:set} %1)
If $F$ is a finite set of formulas then 
\,$\alpha \proves F$ \ iff \ for all $f$ in $F$, \,$\alpha \proves f$. 

\item \label{refine3:fact} %2)
If \,$\Ax \imps f$\, then \,$\alpha \proves f$. \ 
Also \,$\varphi \proves f$\, iff \,$\Ax \imps f$. 

\item \label{refine3:fml} %3)
If \,$\Ax \nimps f$\, and \,$\alpha \neq \varphi$\, then \,$\alpha \proves f$\, iff 
there exists $r\rho$ in $R_{sd}[f]$ such that both (\ref{refine3:fml}.\ref{refine3:For}) and 
(\ref{refine3:fml}.\ref{refine3:Against}) hold. 
	\begin{compactenum}[\ref{refine3:fml}.1)]
	\item \label{refine3:For} %3.1)
	$\alpha \proves A(r\rho)$. 

	\item \label{refine3:Against} %3.2)
	For each $s\sigma$ in $R[\neg f]$ either 
	(\ref{refine3:fml}.\ref{refine3:Against}.\ref{refine3:TeamDefeat}) or 
	(\ref{refine3:fml}.\ref{refine3:Against}.\ref{refine3:Disabled}) holds. 
		\begin{compactenum}[\ref{refine3:fml}.\ref{refine3:Against}.1)]
		\item \label{refine3:TeamDefeat} %3.2.1
		There exists $t\tau$ in \,$R_{sd}[f]_{>s\sigma}$\, such that 
		\,$\alpha \proves A(t\tau)$. 

		\item \label{refine3:Disabled} %3.2.2
		$s\sigma$ is disabled by $\alpha$. 
		\end{compactenum}
	\end{compactenum}
\end{compactenum}
\end{Refine} %4.11

The relation $\proves$ offers only two possibilities, $\proves$ and $\nproves$. 
This is not sufficient for disabling, so we introduce the idea of a proof function $P$. 
The range of $P$ is the set of proof values $\{+1, -1\}$. 
Roughly a proof value of $+1$ means that a formula, or set of formulas, has been proved. 
A proof value of $-1$ means that a formula, or set of formulas, has been disproved; 
that is, it has been shown that there is no proof. 
This is just what is needed for disabling, and gives our fourth refinement. 

%Refinement~4.12 
\begin{Refine} \label{Refine:ProofFunction1} 
%\raggedright \parindent = 1.2em
Let \,$(\Ax,R,>)$\, be a plausible-description, $\alpha$ be a proof algorithm, $f$ be a formula, 
and $F$ be a finite set of formulas. 
\begin{compactenum}[1)]
\item \label{refine4:set} %1)
$P(\alpha,F) = +1$ \ iff \ for all $f$ in $F$, \,$P(\alpha,f) = +1$. \nl 
$P(\alpha,F) = -1$ \ iff \ there exists $f$ in $F$ such that \,$P(\alpha,f) = -1$. 

\item \label{refine4:phi} %2)
$P(\varphi,f) = +1$\, iff \,$\Ax \imps f$. \nl 
$P(\varphi,f) = -1$\, iff \,$\Ax \nimps f$. 

\item \label{refine4:fact} %3)
If \,$\Ax \imps f$\, then \,$P(\alpha,f) = +1$. 

\item \label{refine4:P=+1} %4)
If \,$\Ax \nimps f$\, and \,$\alpha \neq \varphi$\, then \,$P(\alpha,f) = +1$\, iff \nl 
there exists $r\rho$ in $R_{sd}[f]$ such that both 
(\ref{refine4:P=+1}.\ref{refine4:Pset=+1}) and (\ref{refine4:P=+1}.\ref{refine4:Against}) hold. 
	\begin{compactenum}[\ref{refine4:P=+1}.1)]
	\item \label{refine4:Pset=+1} %4.1)
	$P(\alpha,A(r\rho)) = +1$. 

	\item \label{refine4:Against} %4.2)
	For each $s\sigma$ in $R[\neg f]$ either 
	(\ref{refine4:P=+1}.\ref{refine4:Against}.\ref{refine4:TeamDefeat}) or 
	(\ref{refine4:P=+1}.\ref{refine4:Against}.\ref{refine4:Disabled}) holds. 
		\begin{compactenum}[\ref{refine4:P=+1}.\ref{refine4:Against}.1)]
		\item \label{refine4:TeamDefeat} %4.2.1
		There exists $t\tau$ in \,$R_{sd}[f]_{>s\sigma}$\, such that 
		\,$P(\alpha,A(t\tau)) = +1$. 

		\item \label{refine4:Disabled} %4.2.2
		$P(\alpha,A(s\sigma)) = -1$. 
		\end{compactenum}
	\end{compactenum}

\item \label{refine4:P=-1} %5)
If \,$\Ax \nimps f$\, and \,$\alpha \neq \varphi$\, then \,$P(\alpha,f) = -1$\, iff \nl 
for each $r\rho$ in $R_{sd}[f]$, either  
(\ref{refine4:P=-1}.\ref{refine4:Pset=-1}) or (\ref{refine4:P=-1}.\ref{refine4:Against}) holds. 
	\begin{compactenum}[\ref{refine4:P=-1}.1)]
	\item \label{refine4:Pset=-1} %5.1)
	$P(\alpha,A(r\rho)) = -1$. 

	\item \label{refine4:unAgainst} %5.2)
	There exists $s\sigma$ in $R[\neg f]$ such that  
	(\ref{refine4:P=-1}.\ref{refine4:unAgainst}.\ref{refine4:unTeamDefeat}) and  
	(\ref{refine4:P=-1}.\ref{refine4:unAgainst}.\ref{refine4:unDisabled}) both hold. 
		\begin{compactenum}[\ref{refine4:P=-1}.\ref{refine4:unAgainst}.1)]
		\item \label{refine4:unTeamDefeat} %5.2.1
		For all $t\tau$ in \,$R_{sd}[f]_{>s\sigma}$, \,$P(\alpha,A(t\tau)) = -1$. 

		\item \label{refine4:unDisabled} %5.2.2
		$P(\alpha,A(s\sigma)) = +1$. 
		\end{compactenum}
	\end{compactenum}
\end{compactenum}
\end{Refine} %4.12

We now have a refinement in which there are no undefined terms. 
Unfortunately Refinement~\ref{Refine:ProofFunction1} has two failings. 
Apart from the factual proof algorithm $\varphi$, 
there is only one other proof algorithm (denoted by $\alpha$). 
Hence the Many Proof Algorithms Principle, Principle~\ref{Prin:Many Proof Algorithms}, fails.
Also, a proof may get into a loop, and therefore the No Loops Principle, 
Principle~\ref{Prin:No Loops}, fails. 

Before we consider looping, let us invent the other proof algorithms. 
The $\alpha$ in Refinement~\ref{Refine:ProofFunction1}(\ref{refine4:P=+1}.\ref{refine4:Against}.\ref{refine4:Disabled})
evaluates evidence against $f$; and this need not be the same $\alpha$ as in 
(\ref{refine4:P=+1}.\ref{refine4:Pset=+1}) and 
(\ref{refine4:P=+1}.\ref{refine4:Against}.\ref{refine4:TeamDefeat}) 
which evaluates evidence for $f$. 
To avoid confusion let us call the $\alpha$ in 
(\ref{refine4:P=+1}.\ref{refine4:Against}.\ref{refine4:Disabled}), $\alpha'$. 
Replacing $\alpha$ by $\alpha'$ in 
(\ref{refine4:P=+1}.\ref{refine4:Against}.\ref{refine4:Disabled}) 
creates the need to decide what $(\alpha')'$ is. 
Let us simplify $(\alpha')'$ to $\alpha''$. 
Some obvious choices are: \,$\alpha'' = \alpha$, or \,$\alpha'' = \alpha'$, 
or $\alpha''$ is some other proof algorithm. 
The third choice postpones and complicates the choice that must eventually be made. 
Experimentation shows that the second choice has some properties that we would rather avoid. 
So we let \,$\alpha'' = \alpha$. 
We are not really concerned with any primed algorithm as they only assist 
with the definition of their non-primed co-algorithm. 

Another change that can be made is to the set $R[\neg f]$ of rule-instances that 
a proof algorithm regards as evidence against $f$. 
Let $\Foe(\alpha,f)$ denote the set of rule-instances that $\alpha$ regards 
as the evidence against $f$. 

These ideas give us our fifth refinement. 

%Refinement~4.13 
\begin{Refine} \label{Refine:ProofFunction2} 
%\raggedright \parindent = 1.2em
Let \,$(\Ax,R,>)$\, be a plausible-description, $\alpha$ be a proof algorithm, $f$ be a formula, 
and $F$ be a finite set of formulas. 
\begin{compactenum}[1)]
\item \label{refine5:set} %1)
$P(\alpha,F) = +1$ \ iff \ for all $f$ in $F$, \,$P(\alpha,f) = +1$. \nl 
$P(\alpha,F) = -1$ \ iff \ there exists $f$ in $F$ such that \,$P(\alpha,f) = -1$. 

\item \label{refine5:phi} %2)
$P(\varphi,f) = +1$\, iff \,$\Ax \imps f$. \nl 
$P(\varphi,f) = -1$\, iff \,$\Ax \nimps f$. 

\item \label{refine5:fact} %3)
If \,$\Ax \imps f$\, then \,$P(\alpha,f) = +1$. 

\item \label{refine5:P=+1} %4)
If \,$\Ax \nimps f$\, and \,$\alpha \neq \varphi$\, then \,$P(\alpha,f) = +1$\, iff \nl 
there exists $r\rho$ in $R_{sd}[f]$ such that both 
(\ref{refine5:P=+1}.\ref{refine5:Pset=+1}) and (\ref{refine5:P=+1}.\ref{refine5:Against}) hold. 
	\begin{compactenum}[\ref{refine5:P=+1}.1)]
	\item \label{refine5:Pset=+1} %4.1)
	$P(\alpha,A(r\rho)) = +1$. 

	\item \label{refine5:Against} %4.2)
	For each $s\sigma$ in \,$\Foe(\alpha,f)$\, either 
	(\ref{refine5:P=+1}.\ref{refine5:Against}.\ref{refine5:TeamDefeat}) or 
	(\ref{refine5:P=+1}.\ref{refine5:Against}.\ref{refine5:Disabled}) holds. 
		\begin{compactenum}[\ref{refine5:P=+1}.\ref{refine5:Against}.1)]
		\item \label{refine5:TeamDefeat} %4.2.1
		There exists $t\tau$ in \,$R_{sd}[f]_{>s\sigma}$\, such that 
		\,$P(\alpha,A(t\tau)) = +1$. 

		\item \label{refine5:Disabled} %4.2.2
		$P(\alpha',A(s\sigma)) = -1$. 
		\end{compactenum}
	\end{compactenum}

\item \label{refine5:P=-1} %5)
If \,$\Ax \nimps f$\, and \,$\alpha \neq \varphi$\, then \,$P(\alpha,f) = -1$\, iff \nl 
for each $r\rho$ in $R_{sd}[f]$, either  
(\ref{refine5:P=-1}.\ref{refine5:Pset=-1}) or (\ref{refine5:P=-1}.\ref{refine5:Against}) holds. 
	\begin{compactenum}[\ref{refine5:P=-1}.1)]
	\item \label{refine5:Pset=-1} %5.1)
	$P(\alpha,A(r\rho)) = -1$. 

	\item \label{refine5:unAgainst} %5.2)
	There exists $s\sigma$ in \,$\Foe(\alpha,f)$\, such that  
	(\ref{refine5:P=-1}.\ref{refine5:unAgainst}.\ref{refine5:unTeamDefeat}) and  
	(\ref{refine5:P=-1}.\ref{refine5:unAgainst}.\ref{refine5:unDisabled}) both hold. 
		\begin{compactenum}[\ref{refine5:P=-1}.\ref{refine5:unAgainst}.1)]
		\item \label{refine5:unTeamDefeat} %5.2.1
		For all $t\tau$ in \,$R_{sd}[f]_{>s\sigma}$, \,$P(\alpha,A(t\tau)) = -1$. 

		\item \label{refine5:unDisabled} %5.2.2
		$P(\alpha',A(s\sigma)) = +1$. 
		\end{compactenum}
	\end{compactenum}
\end{compactenum}
\end{Refine} %4.13

Our first non-factual proof algorithm, $\beta$, is created by replacing each $\alpha$ in 
Refinement~\ref{Refine:ProofFunction2} with $\beta$, defining \,$\beta' = \beta$, and 
\,$\Foe(\beta,f) = \Foe(\beta',f) = R[\neg f]$. 
It turns out that $\beta$ is ambiguity blocking (b for beta and blocking). 

If we do not identify $\beta$ with $\beta'$ then surprisingly we get a different ambiguity blocking 
algorithm $\theta$ (th for theta and thwarting) which does not do `best bet' reasoning. 
We define $\theta$ by replacing each $\alpha$ in Refinement~\ref{Refine:ProofFunction2} with 
$\theta$, and defining \,$\Foe(\theta,f) = \Foe(\theta',f) = R[\neg f]$. 
We define $\theta'$ by replacing each $\alpha$ in Refinement~\ref{Refine:ProofFunction2} with 
$\theta'$. 
(Recall that \,$\theta'' = \theta$.) 

Our next algorithm, $\pi$, turns out to be ambiguity propagating (p for pi and propagating). 
We want to make $\pi$ as strong as possible; that is, 
$\pi$ proves $f$ if there is no evidence against $f$. 
This can be done by making its co-algorithm $\pi'$ 
ignore all evidence against $f$; so \,$\Foe(\pi',f) = \{\}$. 
This is the only change to Refinement~\ref{Refine:ProofFunction2}. 
Explicitly, let \,$\Foe(\pi,f) = R[\neg f]$. 
Let $\pi$ be defined by replacing each $\alpha$ in 
Refinement~\ref{Refine:ProofFunction2} with $\pi$. 
Let $\pi'$ be defined by replacing each $\alpha$ in 
Refinement~\ref{Refine:ProofFunction2} with $\pi'$. 
(Recall that \,$\pi'' = \pi$.) 

Our last algorithm, $\psi$, turns out to be ambiguity propagating (p for psi and propagating). 
We want to make $\psi$ weaker than $\pi$. 
This can be done by making its co-algorithm $\psi'$ regard only those rules 
that imply $\neg f$ and are superior to $r\rho$ as evidence against $f$. 
This requires $\Foe$ to have 3 rather than 2 arguments.  
So let \,$\Foe(\psi',f,r\rho) = \{s\sigma \e R[\neg f] : s\sigma > r\rho\} = R[\neg f]_{>r\rho}$. 
This is the only change we make to Refinement~\ref{Refine:ProofFunction2}. 
Explicitly, let \,$\Foe(\psi,f) = R[\neg f]$. 
Let $\psi$ be defined by replacing each $\alpha$ in 
Refinement~\ref{Refine:ProofFunction2} with $\psi$. 
Let $\psi'$ be defined by replacing each $\alpha$ in 
Refinement~\ref{Refine:ProofFunction2} with $\psi'$. 
(Recall that \,$\psi'' = \psi$.) 

To emphasise that we are only interested in the non-primed proof algorithms 
we note that there are examples in which both $\pi'$ and $\psi'$ can 
prove both $f$ and $\neg f$. 
This is fine as both $\pi'$ and $\psi'$ only assess the evidence against $f$, 
rather than try to defeasibly justify accepting $f$, as the non-primed algorithms do. 

%Definition~4.14 
\begin{Defn} \label{Defn:Alg,co-algorithms} 
%\raggedright \parindent = 1.2em
The set, $\Alg$, of names of the eight proof algorithms is defined by \nl
\,$\Alg = \{\varphi,\pi,\psi,\theta,\theta',\beta,\psi',\pi'\}$. 
Define \,$\varphi' = \varphi$\, and \,$\beta' = \beta$. 
Suppose that \,$\alpha \e \Alg$\, and define \,$(\alpha')' = \alpha'' = \alpha$. 
The \defn{co-algorithm} of $\alpha$ is $\alpha'$ and 
the \defn{co-algorithm} of $\alpha'$ is $\alpha$. 
\end{Defn} 

For uniformity we shall regard $\Foe$ as having 3 arguments. 

%Definition~4.15 
\begin{Defn} \label{Defn:Foe(.,.,.)} 
%\raggedright \parindent = 1.2em
Suppose $(\Ax,R,>)$ is a plausible-description, $f$ is a formula, and \,$r\rho \e R\Sigma$. 
\begin{compactenum}[1)]
\item %1) 
	If \,$\alpha \e \{\pi,\psi,\theta,\theta',\beta\}$\, then \,$\Foe(\alpha,f,r\rho) = R[\neg f]$. 
\item %2) 
	If \,$\alpha \e \{\varphi,\pi'\}$\, then \,$\Foe(\alpha,f,r\rho) = \{\}$. 
\item %3) 
	$\Foe(\psi',f,r\rho) = \{s\sigma \e R[\neg f] : s\sigma > r\rho\} = R[\neg f]_{>r\rho}$. 
\end{compactenum}
\end{Defn} 

Finally, let us consider looping. 
To prove $f$ we use $\alpha$ and a rule-instance $r\rho$. 
While proving $f$ we may have to prove other formulas. 
During a proof of one of these other formulas, 
if we choose to use $\alpha$ and $r\rho$ again then 
we will be in a loop and so this choice should fail. 
To prevent such a looping choice we need to record that 
$\alpha$ and $r\rho$ have been used previously. 
We shall call such a record of used algorithms and rule-instances a history. 
Its formal definition follows. 

%Definition~4.16 
\begin{Defn} \label{Defn:history} 
%\raggedright \parindent = 1.2em
Suppose $(\Ax,R,>)$ is a plausible-description and \,$\alpha \e \Alg$. 
Recall that $\Sigma$ is the set of all substitutions. 
Define \,$\alpha R\Sigma = \{\alpha r\sigma : r\sigma \e R\Sigma\}$. 
Then $H$ is an \defn{$\alpha$-history} iff $H$ is a finite sequence of elements of 
\,$\alpha R\Sigma \cup \alpha' R\Sigma$\, that has no repeated elements. 
$H$ is a \defn{history} iff there is a proof algorithm $\alpha$ such that $H$ is an $\alpha$-history. 
\end{Defn} 

Unfortunately using a history complicates Refinement~\ref{Refine:ProofFunction2} 
because we now no longer have just an algorithm proving a formula, 
but an algorithm and a history proving a formula. 
Therefore in (\ref{refine5:set}), (\ref{refine5:phi}), (\ref{refine5:fact}), (\ref{refine5:P=+1}), 
and (\ref{refine5:P=-1}), \,$P(\alpha,x)$\, becomes \,$P(\alpha,H,x)$. 

But in (\ref{refine5:P=+1}.\ref{refine5:Pset=+1}) $\alpha$ and $r\rho$ have now been used 
so $H$ must be updated to \,$H$+$\alpha r\rho$\, and 
therefore \,$P(\alpha,H,A(r\rho))$\, becomes \,$P(\alpha,H$+$\alpha r\rho,A(r\rho))$. 
Also to prevent looping we must have \,$\alpha r\rho \nte H$. 
In (\ref{refine5:P=+1}.\ref{refine5:Against}.\ref{refine5:TeamDefeat}) 
$\alpha$ and $t\tau$ have been used so $H$ must be updated to \,$H$+$\alpha t\tau$\, and 
hence \,$P(\alpha,H,A(t\tau))$\, becomes \,$P(\alpha,H$+$\alpha t\tau,A(t\tau))$. 
Also to prevent looping we must have \,$\alpha t\tau \nte H$. 
In (\ref{refine5:P=+1}.\ref{refine5:Against}.\ref{refine5:Disabled}) 
$\alpha'$ and $s\sigma$ have been used and so $H$ must be updated to 
\,$H$+$\alpha' s\sigma$\, and hence 
\,$P(\alpha',H,A(s\sigma))$\, becomes \,$P(\alpha',H$+$\alpha' s\sigma,A(s\sigma))$. 
Also to prevent looping we must have \,$\alpha' s\sigma \nte H$. 

Similarly, in (\ref{refine5:P=-1}) we must make changes that correspond to 
the changes in (\ref{refine5:P=+1}). 

Incorporating these changes into Refinement~\ref{Refine:ProofFunction2} gives 
our formal definition of the proof algorithms and proof function $P$. 
(Although $P$ will be a partial function, we shall nevertheless refer to it as a function.) 

%Definition~4.17 
\begin{Defn} \label{Defn:P, algs} 
%\raggedright \parindent = 1.2em
Suppose \,$\calD = (\Ax,R,>)$\, is a plausible-description, $\alpha \e \Alg$, 
$H$ is an $\alpha$-history, $f$ is a formula, and $F$ be a finite set of formulas. 
The \defn{proof function} for $\calD$ is denoted by $P$. 
The range of $P$ is $\{+1, -1\}$. 
The source of $P$ is \,$\{(\alpha,H,x) : \alpha \e \Alg, H$ is an $\alpha$-history, and 
$x$ is either a formula or a set of formulas$\}$. 
\begin{compactenum}[P1)]
\item \label{Pset} %P1)
$P(\alpha,H,F) = +1$ \ iff \ for all $f$ in $F$, \,$P(\alpha,H,f) = +1$. \nl 
$P(\alpha,H,F) = -1$ \ iff \ there exists $f$ in $F$ such that \,$P(\alpha,H,f) = -1$. 

\item \label{Pphi} %P2)
If $H$ is a $\varphi$-history then \,$P(\varphi,H,f) = +1$\, iff \,$\Ax \imps f$. \nl 
If $H$ is a $\varphi$-history then \,$P(\varphi,H,f) = -1$\, iff \,$\Ax \nimps f$. 

\item \label{Pfact} %P3)
If \,$\Ax \imps f$\, then \,$P(\alpha,H,f) = +1$. 

\item \label{P=+1} %P4)
If \,$\Ax \nimps f$\, and \,$\alpha \neq \varphi$\, then \,$P(\alpha,H,f) = +1$\, iff \nl 
there exists $r\rho$ in $R_{sd}[f]$ such that both 
(P\ref{P=+1}.\ref{Pset=+1}) and (P\ref{P=+1}.\ref{Against}) hold. 
	\begin{compactenum}[P\ref{P=+1}.1)]
	\item \label{Pset=+1} %P4.1)
	$\alpha r\rho \nte H$\, and \,$P(\alpha,H$+$\alpha r\rho,A(r\rho)) = +1$. 

	\item \label{Against} %P4.2)
	For each $s\sigma$ in $\Foe(\alpha,f,r\rho)$ either 
	(P\ref{P=+1}.\ref{Against}.\ref{TeamDefeat}) or 
	(P\ref{P=+1}.\ref{Against}.\ref{Disabled}) holds. 
		\begin{compactenum}[P\ref{P=+1}.\ref{Against}.1)]
		\item \label{TeamDefeat} %P4.2.1
		There exists $t\tau$ in \,$R_{sd}[f]_{>s\sigma}$\, such that 
		\,$\alpha t\tau \nte H$\, and \nl 
		$P(\alpha,H$+$\alpha t\tau,A(t\tau)) = +1$. 

		\item \label{Disabled} %P4.2.2
		$\alpha' s\sigma \nte H$\, and \,$P(\alpha',H$+$\alpha' s\sigma,A(s\sigma)) = -1$. 
		\end{compactenum}
	\end{compactenum}

\item \label{P=-1} %P5)
If \,$\Ax \nimps f$\, and \,$\alpha \neq \varphi$\, then \,$P(\alpha,H,f) = -1$\, iff \nl 
for each $r\rho$ in $R_{sd}[f]$, either  
(P\ref{P=-1}.\ref{Pset=-1}) or (P\ref{P=-1}.\ref{unAgainst}) holds. 
	\begin{compactenum}[P\ref{P=-1}.1)]
	\item \label{Pset=-1} %P5.1)
	$\alpha r\rho \e H$\, or \,$P(\alpha,H$+$\alpha r\rho,A(r\rho)) = -1$. 
	
	\item \label{unAgainst} %P5.2)
	There exists $s\sigma$ in $\Foe(\alpha,f,r\rho)$ such that  
	(P\ref{P=-1}.\ref{unAgainst}.\ref{unTeamDefeat}) and  
	(P\ref{P=-1}.\ref{unAgainst}.\ref{unDisabled}) both hold. 
		\begin{compactenum}[P\ref{P=-1}.\ref{unAgainst}.1)]
		\item \label{unTeamDefeat} %P5.2.1
		For all $t\tau$ in \,$R_{sd}[f]_{>s\sigma}$, \,$\alpha t\tau \e H$\, or 
		\,$P(\alpha,H$+$\alpha t\tau,A(t\tau)) = -1$. 

		\item \label{unDisabled} %P5.2.2
		$\alpha' s\sigma \e H$\, or \,$P(\alpha',H$+$\alpha' s\sigma,A(s\sigma)) = +1$. 
		\end{compactenum}
	\end{compactenum}
\end{compactenum}
\end{Defn} %4.17
(In the above definition of $P$, since the range of $P$ is $\{+1, -1\}$, we should really write 
``$P(x,y,z) = +1$ and $P(x,y,z) \neq -1$'' instead of just ``$P(x,y,z) = +1$''. 
Also we should really write ``$P(x,y,z) = -1$ and $P(x,y,z) \neq +1$'' instead of just 
``$P(x,y,z) = -1$''. 
It is then straightforward to show that $P$ is a (well-defined partial) function. 
However, since we know that $P$ is meant to be a (partial) function, 
this added complication can be implicit rather than explicit.) 

We can now define the relation, $\proves$, which is sometimes useful. 

%Definition~4.18
\begin{Defn} \label{Defn:alpha-provable, calD(alpha)} 
%\raggedright \parindent = 1.2em
Suppose $\calD$ is a plausible-description, \,$\alpha \e \Alg$, $H$ is an $\alpha$-history, and 
$x$ is either a formula or a finite set of formulas. 
\begin{compactenum}[1)]
\item %1)
	We say $x$ is \defn{$\alpha$-provable}, or $\alpha$ proves $x$, 
	and write \,$\alpha \proves x$, \,iff \,$P(\alpha,(),x) = +1$. 
\item %2)
	$(\alpha,H) \proves x$\, iff \,$P(\alpha,H,x) = +1$. 
\item %3) 
	$(\alpha,H) \nproves x$\, means \,not[$(\alpha,H) \proves x$], and 
	\,$\alpha \nproves x$\, means \,not[$\alpha \proves x$]. 
\end{compactenum}
\end{Defn} 
We explicitly note that \,$(\alpha,H) \nproves x$\, does not imply \,$P(\alpha,H,x) = -1$. 

We can now formally define what we mean by Plausible Logic. 

%Definition~4.19
\begin{Defn} \label{Defn:plausible logic} 
%\raggedright \parindent = 1.2em
A \defn{Plausible Logic} is a plausible-description and its proof function.
\end{Defn} 

Definition~\ref{Defn:P, algs} is too cumbersome for manual derivations. 
So we shall define two simplifying partial functions: 
$\For$ (evidence for), and \,$\Dftd$ (defeated), 
although we shall nevertheless refer to them as functions. 
The definition of $\For$ and $\Dftd$ (Definition~\ref{Defn:For, Dftd}) requires that 
the operations, $\min$, $\max$, and $-$, be defined on any subset of \,$\{+1, -1\}$. 
These arithmetical operations are defined next.  

%Definition~4.20
\begin{Defn} \label{Defn:operations on proof values} 
%\raggedright \parindent = 1.2em
The set of all \defn{proof values} is \,$\{+1, -1\}$. 
Suppose \,$S \!\subseteq\! \{+1, -1\}$. \nl 
\begin{tabular}{l@{\hspace{3em}}l@{\hspace{3em}}l}
$\min S = -1$\, iff \,$-1 \e S$. & $\min S = +1$\, iff \,$S \!\subseteq\! \{+1\}$. & 
$-$\,$-1 = +1$. \\
$\max S = +1$\, iff \,$+1 \e S$. & $\max S = -1$\, iff \,$S \!\subseteq\! \{-1\}$. & 
$-$\,$+1 = -1$. 
\end{tabular}
\end{Defn} 

%Definition~4.21
\begin{Defn} \label{Defn:For, Dftd} 
%\raggedright \parindent = 1.2em

Suppose $(\Ax,R,>)$ is any plausible-description, \,$\alpha \e \Alg$, 
$H$ is any $\alpha$-history, $f$ is any formula, and $F$ is any finite set of formulas. 
The partial functions $\For$ and $\Dftd$ are defined by (\ref{range, domains}) to (\ref{Dftd=-1}). 
\begin{compactenum}[1)]

\item \label{range, domains} %1) 
The range of $\For$ and $\Dftd$ is the set of proof values $\{+1, -1\}$. \nl 
The source of $\For$ is \,$\{(\alpha,H,f,r\rho) : \alpha \in \Alg\!-\!\{\varphi\}, 
H$ is an $\alpha$-history, $f$ is a formula, \,$\Ax \nimps f$, and \,$r\rho \e R_{sd}[f]\}$. \nl  
The source of $\Dftd$ is \,$\{(\alpha,H,f,r\rho,s\sigma) : \alpha \in \Alg\!-\!\{\varphi\}, 
H$ is an $\alpha$-history, $f$ is a formula, \,$\Ax \nimps f$, \,$r\rho \e R_{sd}[f]$, and 
\,$s\sigma \e \Foe(\alpha,f,r\rho)\}$. 

\item \label{For=+1} %2) 
Suppose \,$\Ax \nimps f$, \,$\alpha \neq \varphi$, and \,$r\rho \e R_{sd}[f]$.  \nl 
$\For(\alpha,H,f,r\rho) = +1$ \ iff \ $\alpha r\rho \nte H$\, and 
\,$P(\alpha,H$+$\alpha r\rho,A(r\rho)) = +1$\, and \nl 
for all $s\sigma$ in $\Foe(\alpha,f,r\rho)$, $\Dftd(\alpha,H,f,r\rho,s\sigma) = +1$. 

\item \label{For=-1} %3) 
Suppose \,$\Ax \nimps f$, \,$\alpha \neq \varphi$, and \,$r\rho \e R_{sd}[f]$.  \nl 
$\For(\alpha,H,f,r\rho) = -1$ \ iff \ $\alpha r\rho \in H$\, or 
\,$P(\alpha,H$+$\alpha r\rho,A(r\rho)) = -1$\, or \nl 
there exists $s\sigma$ in $\Foe(\alpha,f,r\rho)$ such that \,$\Dftd(\alpha,H,f,r\rho,s\sigma) = -1$. 

\item \label{Dftd=+1} %4) 
Suppose \,$\Ax \nimps f$, \,$\alpha \neq \varphi$, \,$r\rho \e R_{sd}[f]$, and 
\,$s\sigma \e \Foe(\alpha,f,r\rho)$. \nl 
$\Dftd(\alpha,H,f,r\rho,s\sigma) = +1$ \ iff 
either (\ref{Dftd=+1}.1) or (\ref{Dftd=+1}.2). \nl 
(\ref{Dftd=+1}.1) There exists $t\tau$ in $R_{sd}[f]_{>s\sigma}$ such that 
\,$\alpha t\tau \nte H$\, and %\nl 
\,$P(\alpha,H$+$\alpha t\tau,A(t\tau)) = +1$.  \nl 
(\ref{Dftd=+1}.2) $\alpha's\sigma \nte H$\, and 
\,$P(\alpha', H$+$\alpha's\sigma,A(s\sigma)) = -1$. 

\item \label{Dftd=-1} %5) 
Suppose \,$\Ax \nimps f$, \,$\alpha \neq \varphi$, \,$r\rho \e R_{sd}[f]$, and 
\,$s\sigma \e \Foe(\alpha,f,r\rho)$. \nl 
$\Dftd(\alpha,H,f,r\rho,s\sigma) = -1$ \ iff 
(\ref{Dftd=-1}.1) and (\ref{Dftd=-1}.2). \nl 
(\ref{Dftd=-1}.1) for all $t\tau$ in $R_{sd}[f]_{>s\sigma}$, 
either \,$\alpha t\tau \in H$\, or \,$P(\alpha,H$+$\alpha t\tau,A(t\tau)) = -1$. \nl 
(\ref{Dftd=-1}.2) either \,$\alpha's\sigma \in H$\, or 
\,$P(\alpha', H$+$\alpha's\sigma,A(s\sigma)) = +1$. 

\end{compactenum}
Define $E$ to be an \defn{evaluation function} iff \,$E \e \{P, \For, \Dftd\}$. 
\end{Defn} %4.21
(A remark similar to the parenthetical remark at the end of Definition~\ref{Defn:P, algs} about 
implicit and explicit complications applies to the above definition of $\For$ and $\Dftd$.) 

The following theorem provides a characterisation of $P$, $\For$, and $\Dftd$, 
using $\max$ and $\min$, (hence the prefix M), that is easier to use than their definitions. 

%%%%%%%%%%%%%%%%%%%%%%%%%%%%%%%%%%%%%%%%%%%
%Theorem~4.22
\begin{Thm} \label{Thm:P, For, Dftd} 
%\raggedright \parindent = 1.2em
Suppose $(\Ax,R,>)$ is any plausible-description, \,$\alpha \e \Alg$, 
$H$ is any $\alpha$-history, $f$ is any formula, and $F$ is any finite set of formulas. 
\begin{compactenum}[M1)]

\item \label{Mfml} %M1) 
If \,$\Ax \nimps f$\, and \,$\alpha \neq \varphi$\, then 
\,$P(\alpha,H,f) = \max\{\For(\alpha,H,f,r\rho) : r\rho \e R_{sd}[f]\}$. 

\item \label{MFor} %M2) 
If \,$\Ax \nimps f$, \,$\alpha \neq \varphi$, and \,$r\rho \e R_{sd}[f]$\, then 
\,$\For(\alpha,H,f,r\rho) = \min\{x,n\}$, where \nl
\,$x = \max\{P(\alpha,H$+$\alpha r\rho,A(r\rho)) : \alpha r\rho \nte H\}$,  and \nl 
\,$n = \min\{\Dftd(\alpha,H,f,r\rho,s\sigma) : s\sigma \e \Foe(\alpha,f,r\rho)\}$. 

\item \label{MDftd} %M3) 
If \,$\Ax \nimps f$, \,$\alpha \neq \varphi$, \,$r\rho \e R_{sd}[f]$, and 
\,$s\sigma \e \Foe(\alpha,f,r\rho)$\, then \nl 
\,$\Dftd(\alpha,H,f,r\rho,s\sigma) = \max(X \cup Y)$, where \nl 
\,$X = \{P(\alpha,H$+$\alpha t\tau,A(t\tau)) : t\tau \e R_{sd}[f]_{>s\sigma}$\, and 
\,$\alpha t\tau \nte H\}$, and \nl 
\,$Y = \{-P(\alpha', H$+$\alpha's\sigma, A(s\sigma)) : \alpha's\sigma \nte H\}$. 

\item \label{Mset} %M4)  
$P(\alpha,H,F) = \min\{P(\alpha,H,f) : f \e F\}$. 

\end{compactenum} 
\end{Thm} %4.22
%%%%%%%%%%%%%%%%%%%%%%%%%%%%%%%%%%%%%%%

In mathematical logic a proof is a mathematical object; usually a sequence of formulas. 
In plausible logic a proof will be a rooted acyclic digraph (rad) that 
records the application of Theorem~\ref{Thm:P, For, Dftd}. 
In M\ref{Mset}, we can regard \,$\{(\alpha,H,f) : f \e F\}$\, as 
the set of children of $(\alpha,H,F)$.
A similar remark applies to M\ref{Mfml}, M\ref{MFor}, and M\ref{MDftd}. 
So the application of Theorem~\ref{Thm:P, For, Dftd} can be regarded as growing a rad. 

Typically we want to show that $\alpha$ either can, or cannot, prove a formula $f$ by 
applying Theorem~\ref{Thm:P, For, Dftd} to evaluate $P(\alpha,(),f)$. 
This will generate a subrad of a full evaluation rad, which we now define. 

%Definition~4.23
\begin{Defn} \label{Defn:full evaluation rad} 
%\raggedright \parindent = 1.2em
Suppose \,$\calD = (\Ax,R,>)$\, is any plausible-description, \,$\alpha \e \Alg$, 
$H$ is any $\alpha$-history, $F$ is a finite set of formulas, $f$ is any formula, and 
\,$\{r\rho, s\sigma\} \subseteq R\Sigma$. 
A \defn{full evaluation rad} of $\calD$, $T$, is a rooted acyclic digraph (rad) constructed 
as follows. 
Each node of $T$ has one of the following forms: $(\alpha,H,f)$, $(\alpha,H,f,r\rho)$, 
$(\alpha,H,f,r\rho,s\sigma)$, or $(\alpha,H,F)$. 
If $(x)$ is a node then it is convenient to simplify notation by defining \,$\Ch(x) = \Ch((x))$, 
the set of children of $(x)$. 
\begin{compactenum}[T1)]

\item \label{TPhi} %T1) 
If $(\alpha,H,f)$ is a node of $T$ and \,$\alpha = \varphi$\, then \,$\Ch(\alpha,H,f) = \{\}$. 

\item \label{Tfact} %T2)
If $(\alpha,H,f)$ is a node of $T$ and \,$\Ax \imps f$\, then \,$\Ch(\alpha,H,f) = \{\}$. 

\item \label{Tfml} %T3)
If $(\alpha,H,f)$ is a node of $T$ and \,$\Ax \nimps f$\, and \,$\alpha \neq \varphi$\, then \nl 
\,$\Ch(\alpha,H,f) = \{(\alpha,H,f,r\rho) : r\rho \e R_{sd}[f]\}$. 

\item \label{TFor} %T4)
If $(\alpha,H,f,r\rho)$ is a node of $T$, \,$\Ax \nimps f$, \,$\alpha \neq \varphi$, and 
\,$r\rho \e R_{sd}[f]$\, then \nl 
\,$\Ch(\alpha,H,f,r\rho) = \{(\alpha,H$+$\alpha r\rho,A(r\rho)) : \alpha r\rho \nte H\}$ $\cup$ \nl 
\hspace*{7.8em}$\{(\alpha,H,f,r\rho,s\sigma) : s\sigma \e \Foe(\alpha,f,r\rho)\}$. 

\item \label{TDftd} %T5)
If $(\alpha,H,f,r\rho,s\sigma)$ is a node of $T$, \,$\Ax \nimps f$, \,$\alpha \neq \varphi$, 
\,$r\rho \e R_{sd}[f]$, and \nl 
\,$s\sigma \e \Foe(\alpha,f,r\rho)$\, then \nl 
\,$\Ch(\alpha,H,f,r\rho,s\sigma) = 
\{(\alpha,H$+$\alpha t\tau,A(t\tau)) : t\tau \e R_{sd}[f]_{>s\sigma}$\, and 
\,$\alpha t\tau \nte H\}$ $\cup$ \nl 
\hspace*{9.3em}$\{(\alpha', H$+$\alpha's\sigma, A(s\sigma)) : \alpha's\sigma \nte H\}$. 

\item \label{Tset} %T6)
If $(\alpha,H,F)$ is a node of $T$ then \,$\Ch(\alpha,H,F) = \{(\alpha,H,f) : f \e F\}$. 

\end{compactenum}
\end{Defn} %4.23

Examples show that a full evaluation rad need not be a layered rad, 
and so a node can have more than one parent. 
Examples show that a full evaluation rad can have an infinite path, 
and so the height of the rad is $\omega$. 
However, there are examples of a full evaluation rad in which each path is finite, 
but its height is still $\omega$. 

Roughly a proof in PL will be a subrad of a full evaluation rad $T$ that contains 
just what is needed. 
To decide what is needed, the nodes of $T$ are classified in 
Definition~\ref{Defn:necessary optional irrelevant nodes}. 
But first the following definition is needed. 

%Definition~4.24
\begin{Defn} \label{Defn:Dom(.)} 
%\raggedright \parindent = 1.2em
Suppose \,$\calD = (\Ax,R,>)$\, is any plausible-description, \,$\alpha \e \Alg$, 
$H$ is any $\alpha$-history, $f$ is any formula, $x$ is either a formula or a finite set of formulas, 
and \,$\{r\rho, s\sigma\} \subseteq R\Sigma$. 
\begin{compactenum}[1)]
\item %1)
	$\Dom(P) = \{(\alpha,H,x) : P(\alpha,H,x) \e \{+1,-1\}\}$. 
\item %2)
	$\Dom(\For) = \{(\alpha,H,f,r\rho) : \For(\alpha,H,f,r\rho) \e \{+1,-1\}\}$. 
\item %3)
$\Dom(\Dftd) = \{(\alpha,H,f,r\rho,s\sigma) : \Dftd(\alpha,H,f,r\rho,s\sigma) \e \{+1,-1\}\}$. 
\item %4)
	$\Dom(P,\For,\Dftd) = \Dom(P) \cup \Dom(\For) \cup \Dom(\Dftd)$. 
\end{compactenum}
\end{Defn} %4.24

%Definition~4.25
\begin{Defn} \label{Defn:necessary optional irrelevant nodes} 
%\raggedright \parindent = 1.2em
Let $T$ be a full evaluation rooted acyclic digraph (rad) of a plausible-description $(\Ax, R, >)$. 
Suppose \,$\alpha \e \Alg$, $H$ is an $\alpha$-history, $f$ is a formula, 
$F$ is a finite set of formulas, \,$\Ax \nimps f$, \,$\alpha \neq \varphi$, 
\,$r\rho \e R_{sd}[f]$, and \,$s\sigma \e \Foe(\alpha,f,r\rho)$. 
If \,$p = (\alpha,H,...)$\, is a node of $T$ and \,$E \e \{P,\For,\Dftd\}$\, then define 
\,$E(p) = E(\alpha,H,...)$. 
\begin{compactenum}[1)]
\item  \label{root} %1)
	The root of $T$ is necessary. 
\item \label{(alpha,H,F)} %2)
	If \,$p = (\alpha,H,F)$\, and \,$P(p) = +1$\, then every child of $p$ is necessary for $p$. 
\item  \label{(alpha,H,F),(alpha,H,f)} %3)
Let \,$p = (\alpha,H,F)$\, and \,$c = (\alpha,H,f)$\, be a child of $p$. 
	\begin{compactenum}[\ref{(alpha,H,F),(alpha,H,f)}.1)]
	\item  %3.1)
	If \,$P(p) = -1$\, then $c$ is optional for $p$ iff \,$P(c) = -1$. 
	\item  %3.2)
	If \,$P(p) = -1$\, then $c$ is irrelevant for $p$ iff \,$P(c) \neq -1$. 
	\item  %3.3)
	If \,$p \nte \Dom(P)$\, then $c$ is optional for $p$ iff \,$c \nte \Dom(P)$. 
	\item  %3.4)
	If \,$p \nte \Dom(P)$\, then $c$ is irrelevant for $p$ iff \,$c \e \Dom(P)$. 
	\end{compactenum}
\item  \label{(alpha,H,f)} %4)
	If \,$p = (\alpha,H,f)$\, and \,$P(p) = -1$\, then every child of $p$ is necessary for $p$. 
\item  \label{(alpha,H,f),(alpha,H,f,rrho)} %5)
Let \,$p = (\alpha,H,f)$\, and \,$c = (\alpha,H,f,r\rho)$\, be a child of $p$. 
	\begin{compactenum}[\ref{(alpha,H,f),(alpha,H,f,rrho)}.1)]
	\item  %5.1)
	If \,$P(p) = +1$\, then $c$ is optional for $p$ iff \,$\For(c) = +1$. 
	\item  %5.2)
	If \,$P(p) = +1$\, then $c$ is irrelevant for $p$ iff \,$\For(c) \neq +1$. 
	\item  %5.3)
	If \,$p \nte \Dom(P)$\, then $c$ is optional for $p$ iff \,$c \nte \Dom(\For)$. 
	\item  %5.4)
	If \,$p \nte \Dom(P)$\, then $c$ is irrelevant for $p$ iff \,$c \e \Dom(\For)$. 
	\end{compactenum}
\item  \label{(alpha,H,f,rrho)} %6)
	If \,$p = (\alpha,H,f,r\rho)$\, and \,$\For(p) = +1$\, then 
	every child of $p$ is necessary for $p$. 
\item  \label{(alpha,H,f,rrho),(alpha,H+alpha rrho,A(rrho))} %7)
Let \,$p = (\alpha,H,f,r\rho)$\, and \,$c = (\alpha,H$+$\alpha r\rho,A(r\rho))$\, be a child of $p$. 
	\begin{compactenum}[\ref{(alpha,H,f,rrho),(alpha,H+alpha rrho,A(rrho))}.1)]
	\item  %7.1)
	If \,$\For(p) = -1$\, then $c$ is optional for $p$ iff \,$P(c) = -1$. 
	\item  %7.2)
	If \,$\For(p) = -1$\, then $c$ is irrelevant for $p$ iff \,$P(c) \neq -1$. 
	\item  %7.3)
	If \,$p \nte \Dom(\For)$\, then $c$ is optional for $p$ iff \,$c \nte \Dom(P)$. 
	\item  %7.4)
	If \,$p \nte \Dom(\For)$\, then $c$ is irrelevant for $p$ iff \,$c \e \Dom(P)$. 
	\end{compactenum}
\item  \label{(alpha,H,f,rrho),(alpha,H,f,rrho,ssigma)} %8)
Let \,$p = (\alpha,H,f,r\rho)$\, and \,$c = (\alpha,H,f,r\rho,s\sigma)$\, be a child of $p$. 
	\begin{compactenum}[\ref{(alpha,H,f,rrho),(alpha,H,f,rrho,ssigma)}.1)]
	\item  %8.1)
	If \,$\For(p) = -1$\, then $c$ is optional for $p$ iff \,$\Dftd(c) = -1$. 
	\item  %8.2)
	If \,$\For(p) = -1$\, then $c$ is irrelevant for $p$ iff \,$\Dftd(c) \neq -1$. 
	\item  %8.3)
	If \,$p \nte \Dom(\For)$\, then $c$ is optional for $p$ iff \,$c \nte \Dom(\Dftd)$. 
	\item  %8.4)
	If \,$p \nte \Dom(\For)$\, then $c$ is irrelevant for $p$ iff \,$c \e \Dom(\Dftd)$. 
	\end{compactenum}
\item  \label{(alpha,H,f,rrho,ssigma)} %9)
	If \,$p = (\alpha,H,f,r\rho,s\sigma)$\, and \,$\Dftd(p) = -1$\, then 
	every child of $p$ is necessary for $p$. 
\item \label{(alpha,H,f,rrho,ssigma),(alpha,H+alpha ttau,A(ttau))} %10)
Let \,$p = (\alpha,H,f,r\rho,s\sigma)$\, and \,$c = (\alpha,H$+$\alpha t\tau,A(t\tau))$\, 
be a child of $p$. 
	\begin{compactenum}[\ref{(alpha,H,f,rrho,ssigma),(alpha,H+alpha ttau,A(ttau))}.1)]
	\item  %10.1)
	If \,$\Dftd(p) = +1$\, then $c$ is optional for $p$ iff \,$P(c) = +1$. 
	\item  %10.2)
	If \,$\Dftd(p) = +1$\, then $c$ is irrelevant for $p$ iff \,$P(c) \neq +1$. 
	\item  %10.3)
	If \,$p \nte \Dom(\Dftd)$\, then $c$ is optional for $p$ iff \,$c \nte \Dom(P)$. 
	\item  %10.4)
	If \,$p \nte \Dom(\Dftd)$\, then $c$ is irrelevant for $p$ iff \,$c \e \Dom(P)$. 
	\end{compactenum}
\item \label{(alpha,H,f,rrho,ssigma),(alpha',H+alpha' ssigma,A(ssigma))} %11)
Let \,$p = (\alpha,H,f,r\rho,s\sigma)$\, and \,$c = (\alpha',H$+$\alpha' s\sigma,A(s\sigma))$\, 
be a child of $p$. 
	\begin{compactenum}[\ref{(alpha,H,f,rrho,ssigma),(alpha',H+alpha' ssigma,A(ssigma))}.1)]
	\item  %11.1)
	If \,$\Dftd(p) = +1$\, then $c$ is optional for $p$ iff \,$P(c) = -1$. 
	\item  %11.2)
	If \,$\Dftd(p) = +1$\, then $c$ is irrelevant for $p$ iff \,$P(c) \neq -1$. 
	\item  %11.3)
	If \,$p \nte \Dom(\Dftd)$\, then $c$ is optional for $p$ iff \,$c \nte \Dom(P)$. 
	\item  %11.4)
	If \,$p \nte \Dom(\Dftd)$\, then $c$ is irrelevant for $p$ iff \,$c \e \Dom(P)$. 
	\end{compactenum}
\end{compactenum}
\end{Defn} %4.25
Since a node in $T$ can have 2 or more parents, it is conceivable that a node could have 
different classifications depending on which parent is being considered. 

Now we define a subrad of a full evaluation rad that contains just what is needed. 

%Definition~4.26
\begin{Defn} \label{Defn:evaluation rad} 
%\raggedright \parindent = 1.2em
Let \,$T = (N, A, r)$\, be a full evaluation rad of a plausible-description $\calD$. 
Then $T_e$ is an \defn{evaluation rad} of $\calD$ iff the following 4 conditions all hold. 
\begin{compactenum}[1)]
\item  %1)
	$T_e = (N_e, A_e, r)$. \ (So $T$ and $T_e$ have the same root.) 
\item %2) 
	If \,$p \e N_e$\, and \,$(p,c) \e A$\, and $c$ is necessary for $p$ then \,$(p,c) \e A_e$. 
\item %3)
	Let \,$O(p,T) = \{c : (p,c) \e A$ and $c$ optional for $p\}$. 
	If \,$p \e N_e$\, and \,$O(p,T) \neq \{\}$\, 
	then there is exactly one element of $O(p,T)$, say $c$, such that \,$(p,c) \e A_e$. 
\item %4)
	If \,$p \e N_e$\, and \,$(p,c) \e A$\, and $c$ is irrelevant for $p$ then \,$c \nte N_e$. 	
\end{compactenum}
\end{Defn} 

%Definition~4.27
\begin{Defn} \label{Defn:proof, disproof} 
%\raggedright \parindent = 1.2em
Suppose that $T$ is an evaluation rad of a plausible-description, and that \,$\alpha \e \Alg$. 
Let $x$ be either a formula or a finite set of formulas. 
\begin{compactenum}[1)]
\item  %1)
	$T$ is an \defn{$\alpha$-proof} of $x$ iff the root of $T$ is $(\alpha, (), x)$ and 
	\,$P(\alpha, (), x) = +1$. 
\item %2) 
	$T$ is an \defn{$\alpha$-disproof} of $x$ iff the root of $T$ is $(\alpha, (), x)$ and 
	\,$P(\alpha, (), x) = -1$. 
\end{compactenum}
\end{Defn} %4.27

We closed this section with two important results. 

%%%%%%%%%%%%%%%%%%%%%%%%%%%%%%%%%%%%%%%%%%
%Theorem~4.28
\begin{Thm} \label{Thm:eval rad is infinite path} 
%\raggedright \parindent = 1.2em
Let $\calD$ be a plausible-description and $T$ be an evaluation rad of $\calD$. 
Then the following 3 statements are equivalent. 
\begin{compactenum}[1)]
\item %1)
	The root of $T$ is not in $\Dom(P,\For,\Dftd)$. 
\item %2) 
	Every node of $T$ is not in $\Dom(P,\For,\Dftd)$. 
\item %3) 
	$T$ is an infinite path. 
\end{compactenum}
\end{Thm} %4.28
%%%%%%%%%%%%%%%%%%%%%%%%%%%%%%%%%%%%%%%%%%

%%%%%%%%%%%%%%%%%%%%%%%%%%%%%%%%%%%%%%%%%%
%Theorem~4.29
\begin{Thm} \label{Thm:eval rad, paths finite} 
%\raggedright \parindent = 1.2em
Let $\calD$ be a plausible-description and $T$ be an evaluation rad of $\calD$. 
Then the following 3 statements are equivalent. 
\begin{compactenum}[1)]
\item %1)
	The root of $T$ is in $\Dom(P,\For,\Dftd)$. 
\item %2) 
	Every node of $T$ is in $\Dom(P,\For,\Dftd)$. 
\item %3) 
	Every path in $T$ is finite. 
\end{compactenum}
\end{Thm} %4.29
%%%%%%%%%%%%%%%%%%%%%%%%%%%%%%%%%%%%%%%%%%

%2345678901234567890123456789012345678901234567890123456789012345678901234567890
%Section 5
\section{Properties of Plausible Logic} 
\label{Section:Properties of PL}
Principle~\ref{Prin:Plausibly Conjunctive} (The Plausibly Conjunctive Principle) says 
that ``a plausible proof algorithm must be plausibly conjunctive''. 
For Plausible Logic this means that whenever \,$\alpha \e \{\pi,\psi,\theta,\theta',\beta\}$, and 
\,$\Ax \imps f$, and \,$\alpha \proves g$\, then \,$\alpha \proves \AND\{f,g\}$. 
This is a special case of the following theorem. 

%Theorem~5.1
\begin{Thm} [Plausible Conjunction] 
\label{Thm:Plausible Conjunction}
%\raggedright \parindent = 1.2em
Let $(\Ax,R,>)$ be a plausible-description, and both $f$ and $g$ be formulas. 
Suppose \,$\alpha \e \Alg$\, and $H$ is an $\alpha$-history. 
If \,$\Ax \imps f$\, and \,$P(\alpha,H,g) = +1$\, then \,$P(\alpha,H,\AND \{f,g\}) = +1$. 
\end{Thm}  %EndThm 5.1

Principle~\ref{Prin:Plausible Right Weakening} (The Plausible Right Weakening Principle) says 
that ``a plausible proof algorithm must have the plausible right weakening property''. 
For Plausible Logic this means that if \,$\alpha \e \{\pi,\psi,\theta,\theta',\beta\}$, and 
\,$\alpha \proves f$, and \,$\Ax \cup \{f\} \imps g$, then \,$\alpha \proves g$. 
This is a special case of the following theorem. 

%Theorem~5.2
\begin{Thm} [Right Weakening] 
\label{Thm:Right Weakening}
%\raggedright \parindent = 1.2em
Let $(\Ax,R,>)$ be a plausible-description, and both $f$ and $g$ be formulas. 
Suppose \,$\alpha \e \Alg$\, and $H$ is an $\alpha$-history. 
\begin{compactenum}[1)]
\item \label{Strong Right Weakening} %1)
If \,$P(\alpha,H,f) = +1$\, and \,$\Ax \cup \{f\} \imps g$\, then \,$P(\alpha,H,g) = +1$. %\nl
	[Strong Right Weakening]
\item \label{Right Weakening} %2)
If \,$P(\alpha,H,f) = +1$\, and \,$f \imps g$\, then \,$P(\alpha,H,g) = +1$. 
	[Right Weakening]
\end{compactenum}
\end{Thm}  %EndThm 5.2

In classical logic Modus Ponens holds for material implication. 
That is, if a formula $f$ is provable and $f$ materially implies a formula $g$ then $g$ is provable. 
A similar result holds for strict rules. 

%Theorem~5.3
\begin{Thm} [Modus Ponens] 
\label{Thm:Modus Ponens}
%\raggedright \parindent = 1.2em
Let $(\Ax,R,>)$ be a plausible-description, and both $f$ and $g$ be formulas. 
Suppose \,$\alpha \e \Alg$\, and $H$ is an $\alpha$-history. 
\begin{compactenum}[1)]
\item \label{Modus Ponens for strict rule-instances} %1)
If \,$r\rho \e R_s\Sigma$\, and \,$P(\alpha,H,A(r\rho)) = +1$\, 
then \,$P(\alpha,H,c(r\rho)) = +1$. \nl 
	[Modus Ponens for strict rule-instances]
\item \label{Modus Ponens for strict rules} %2)
If \,$r \e R_s$\, and \,$P(\alpha,H,A(r)) = +1$\, then \,$P(\alpha,H,c(r)) = +1$. \nl 
	[Modus Ponens for strict rules]
\end{compactenum}
\end{Thm}  %EndThm 5.3

For proof algorithms $\alpha$ and $\delta$, we want to know if 
every formula that can be proved using $\alpha$ can also be proved using $\delta$. 
We also want to know if every formula that can be disproved using $\alpha$ can 
be disproved using $\delta$. 
Hence the following notation. 

%Defn 5.4
\begin{Defn} \label{Defn:D(alph), D^-(alph)} 
%\raggedright \parindent = 1.2em
Suppose $\calD$ is a plausible-description, and \,$\alpha \e \Alg$. 
\begin{compactenum}[1)]
\item %1)
	$\calD(\alpha) = \{f \e \Fml : P(\alpha, (),  f) = +1\}$\, 
	is the set of all $\alpha$-provable formulas. 
\item %2)
	$\calD^-\!(\alpha) = \{f \e \Fml : P(\alpha, (),  f) = -1\}$\, 
	is the set of all $\alpha$-disprovable formulas. 
\end{compactenum}
\end{Defn} 

The proof algorithms of Plausible Logic form a linear hierarchy. 

%Theorem~5.5
\begin{Thm} [Proof Algorithm Hierarchy] 
\label{Thm:The proof algorithm hierarchy} 
%\raggedright \parindent = 1.2em
Let $\calD \!=\! (\Ax,R,>)$ be a plausible-description. 
\begin{compactenum}[1)]
\item %1)
	$\calD(\varphi) \subseteq \calD(\pi) \subseteq \calD(\psi) \subseteq \calD(\theta) 
	= \calD(\theta')\subseteq \calD(\beta)\subseteq \calD(\psi') \subseteq \calD(\pi')$. 
\item %2)
	If $>$ is empty then \,$\calD(\varphi) \subseteq \calD(\pi) = \calD(\psi) \subseteq 
	\calD(\theta) = \calD(\theta')\subseteq \calD(\beta) \subseteq \calD(\psi') = \calD(\pi')$. 
\item %3)
	$\calD^-\!(\pi') \subseteq \calD^-\!(\psi') \subseteq \calD^-\!(\beta) \subseteq 
	\calD^-\!(\theta') = \calD^-\!(\theta) \subseteq \calD^-\!(\psi) \subseteq 
	\calD^-\!(\pi) \subseteq \calD^-\!(\varphi)$. 
\item %4)
	If $>$ is empty then \nl
	$\calD^-\!(\pi') = \calD^-\!(\psi') \subseteq \calD^-\!(\beta) \subseteq 
	\calD^-\!(\theta') = \calD^-\!(\theta) \subseteq \calD^-\!(\psi) = 
	\calD^-\!(\pi) \subseteq \calD^-\!(\varphi)$. 
\end{compactenum}
\end{Thm} %EndThm5.5 

The linear hierarchy of proof algorithms provides 5 different 
non-numeric confidence levels or levels of reliability. 
The $\varphi$ confidence level, given by the $\varphi$ algorithm, 
is the highest confidence level. 
Indeed since it only reasons with facts its answers are always true. 
When reasoning with plausible information there are 4 different levels of reliability 
given by the 4 plausible proof algorithms $\pi$, $\psi$, $\theta$, and $\beta$. 
The most reliable algorithm is $\pi$, the next reliable is $\psi$, the next reliable is $\theta$, 
and the least reliable is $\beta$. 
Nevertheless it is still reliable enough for its answers to be acted on. 

Therefore use the algorithm highest in the hierarchy that proves the formula being considered. 
If the $\alpha$ proof algorithm proves a formula $f$, 
then $f$ is usually true with a confidence or reliability level of $\alpha$. 

The failure to both prove and disprove a formula is called coherence. 
The next theorem shows that the proof algorithms of Plausible Logic are coherent. 

%Theorem~5.6
\begin{Thm} [Coherence] 
\label{Thm:Coherence}
%\raggedright \parindent = 1.2em
Suppose $(\Ax,R,>)$ is a plausible-description, \,$\alpha \e \Alg$, $H$ is an $\alpha$-history, 
and $x$ is either a formula or finite set of formulas. 
If \,$P(\alpha, H, x) = +1$\, then \,$P(\alpha, H, x) \neq -1$; and 
if \,$P(\alpha, H, x) = -1$\, then \,$P(\alpha, H, x) \neq +1$. 
Therefore $\alpha$ is coherent. 
\end{Thm}  %EndThm 5.6

The Strong 2-Consistency Principle (Principle~\ref{Prin:Strong 2-Consistency}) says that 
``a plausible proof algorithm must be strongly 2-consistent''. 
The next theorem shows that is principle is satisfied. 

%Theorem~5.7
\begin{Thm} [Strong 2-Consistency] 
\label{Thm:Strong 2-Consistency}
%\raggedright \parindent = 1.2em
Suppose that $(\Ax,R,>)$ is a plausible-description, \,$\alpha \e \{\varphi,\pi,\psi,\theta,\beta\}$, 
$H$ is an $\alpha$-history, and both $f$ and $g$ are any formulas. 
If \,$P(\alpha, H, f) = +1$\, and \,$P(\alpha, H, g) = +1$\, 
then \,$\Ax \cup \{f,g\}$\, is satisfiable. 
Therefore, $\alpha$ is strongly 2-consistent. 
\end{Thm}  %EndThm 5.7

We end this section by considering decisiveness. 

%Definition~5.8 
\begin{Defn} \label{Defn:PLdecisive}  
%\raggedright \parindent = 1.2em
Let $\calD$ be a plausible-description and $\calL$ be the language determined by $\calD$. 
Suppose $f$ is a formula in $\calL$ and $\alpha \e \Alg$. 
Let \,$T = (N, A, p_r)$\, be the full evaluation rad of $\calD$ such that \,$p_r = (\alpha, (), f)$. 
If \,$n \e \ZZ^+$\, define \,$T(n) = (N', A', p_r)$\, to be the subrad of $T$ such that \nl
(i) every path from $p_r$ has at most $n$ nodes, and \nl 
(ii) every path from $p_r$ that has less then $n$ nodes ends with a leaf. 
\begin{compactenum}[1)] 
\item %1) 
	$\alpha$ is \defn{decisive for} $f$ iff there exists $n$ in $\ZZ^+$ such that \nl 
	either it is clear from $T(n)$ that there is an $\alpha$-proof of $f$, \nl 
	or it is clear from $T(n)$ that there is not any $\alpha$-proof of $f$. 
\item %2) 
	$\alpha$ is \defn{finitely decisive for} $f$ iff either \,$P(\alpha, (), f) = +1$\, 
	or \,$P(\alpha, (), f) = -1$. 
\item %3) 
	$\alpha$ is \defn{decisive} iff for all formulas $f$ in $\calL$, $\alpha$ is decisive for $f$. 
\item %4) 
	$\alpha$ is \defn{finitely decisive} iff for all formulas $f$ in $\calL$, 
	$\alpha$ is finitely decisive for $f$. 
\end{compactenum} 
\end{Defn} %5.8

If $(\Ax, R, >)$ is a plausible-description then a widely applicable condition for 
a proof algorithm to be finitely decisive is that $R\Sigma$ be finite; as the following lemma shows. 

%Lemma~5.9
\begin{Lem} 
\label{Lem:finite decisiveness}
%\raggedright \parindent = 1.2em
Suppose \,$\calD = (\Ax,R,>)$\, is any plausible-description, and \,$\alpha \e \Alg$.  
Let $T$ be a full evaluation rad of $\calD$. 
If $R\Sigma$ is finite then $\alpha$ is finitely decisive and $T$ is finite. 
\end{Lem} %EndLem 5.9

Necessary and sufficient conditions for $\alpha$ to be finitely decisive are in 
Theorem~\ref{Thm:eval rad, paths finite}. 

%2345678901234567890123456789012345678901234567890123456789012345678901234567890
%Section 6
\section{A Truth Theory} 
\label{Section:Truth}
Consider the possibilities that could occur when the proof algorithm 
$\alpha$ evaluates the evidence for and against the formula $f$. 

If there is sufficient evidence for both $f$ and $\neg f$ then, 
as far as $\alpha$ is concerned, $f$ is ambiguous and $\neg f$ is ambiguous. 
Therefore both $f$ and $\neg f$ should be assigned the ambiguous truth value $\tv{a}$. 

If there is insufficient evidence for both $f$ and $\neg f$ then 
$\alpha$ does not know enough about $f$ or about $\neg f$. 
Hence both $f$ and $\neg f$ should be assigned the undetermined truth value $\tv{u}$. 

If there is sufficient evidence for $f$ but 
insufficient evidence for $\neg f$ then, as far as $\alpha$ is concerned, 
$f$ is not ambiguous and $\neg f$ is not undetermined. 
So $f$ should be assigned the usually true truth value $\tv{t}$, 
and $\neg f$ should be assigned the usually false truth value $\tv{f}$. 

Since the truth value of a formula, $f$, depends on the proof algorithm, $\alpha$, 
evaluating its evidence, we need a truth function $V$ (for veracity) such that $V(\alpha,f)$ 
is in the set of plausible truth values $\{\tv{a}, \tv{t}, \tv{f}, \tv{u}\}$. 
This is accomplished by the next definition. 

%Definition~6.1
\begin{Defn} \label{Defn:V(.,.)} 
%\raggedright \parindent = 1.2em
Suppose \,$\calD = (\Ax,R,>)$\, is a plausible-description, \,$\alpha \e \Alg$, and 
$f$ is any formula. 
The \defn{ambiguous truth value} is $\tv{a}$, the \defn{usually true truth value} is $\tv{t}$, 
the \defn{usually false truth value} is $\tv{f}$, and 
the \defn{undetermined truth value} is $\tv{u}$. 
The \defn{truth function for} $\calD$, $V$, from \,$\Alg \!\times\! \Fml$\, to 
the \defn{set of plausible truth values} $\{\tv{a}, \tv{t}, \tv{f}, \tv{u}\}$ 
is defined by V\ref{tv{a}} to V\ref{tv{u}}. 
\begin{compactenum}[V1)]
\item \label{tv{a}} %1)
	$V(\alpha,f) = \tv{a}$ \ iff \ $\alpha \proves f$\, and \,$\alpha \proves \neg f$. 
\item \label{tv{t}} %2)
	$V(\alpha,f) = \tv{t}$ \ iff \ $\alpha \proves f$\, and \,$\alpha \nproves \neg f$. 
\item \label{tv{f}} %3)
	$V(\alpha,f) = \tv{f}$ \ iff \ $\alpha \nproves f$\, and \,$\alpha \proves \neg f$. 
\item \label{tv{u}} %4)
	$V(\alpha,f) = \tv{u}$ \ iff \ $\alpha \nproves f$\, and \,$\alpha \nproves \neg f$. 
\end{compactenum}
\end{Defn} %6.1

The Included Middle Principle (Principle~\ref{Prin:Included Middle}) is the following. 
``If a logic for plausible reasoning has a truth theory then 
that truth theory must have at least 3 truth values.'' 
So PL satisfies this principle. 

Also in Subsection~\ref{PLPR:Truth Values} we noted that the closest plausible reasoning 
can get to the usual relationship between conjunction and $\tv{t}$, and 
between disjunction and $\tv{t}$, is the following. 
\begin{compactitem}[$\triangleright$]
\item If \,$V(\alpha, \AND\{f,g\}) = \tv{t}$\, then \,$V(\alpha, f) = \tv{t} = V(\alpha, g)$. 
\item If \,$V(\alpha, f) = \tv{t}$\, or \,$V(\alpha, g) = \tv{t}$\, then 
	\,$V(\alpha, \OR\{f,g\}) = \tv{t}$. 
\end{compactitem}
These two results are a special case of 
Theorem~\ref{Thm:Truth values}(\ref{If V(alpha,AND F) = tv{t} then V(alpha,f) = tv{t}},\ref{If V(alpha,f) = tv{t} then V(alpha,OR F) = tv{t}}) below. 

%Theorem~6.2
\begin{Thm} 
\label{Thm:Truth values} 
%\raggedright \parindent = 1.2em
Suppose $(\Ax,R,>)$ is a plausible-description, \,$\alpha \e \Alg$, $f$ is a formula, 
and $F$ is a finite set of formulas. 
\begin{compactenum}[ \ 1)]
\item \label{V(alpha,notnotf) = V(alpha,f)} %1)
	$V(\alpha,\neg \neg f) = V(\alpha,f)$. 
\item \label{V(alpha,f) = tv{t} iff V(alpha,not f) = tv{f}} %2)
	$V(\alpha,f) = \tv{t}$\, iff \,$V(\alpha,\neg f) = \tv{f}$. 
\item \label{V(alpha,f) = tv{f} iff V(alpha,not f) = tv{t}} %3)
	$V(\alpha,f) = \tv{f}$\, iff \,$V(\alpha,\neg f) = \tv{t}$. 
\item \label{V(alpha,f) = tv{a} iff V(alpha,not f) = tv{a}} %4)
	$V(\alpha,f) = \tv{a}$\, iff \,$V(\alpha,\neg f) = \tv{a}$. 
\item \label{V(alpha,f) = tv{u} iff V(alpha,not f) = tv{u}} %5)
	$V(\alpha,f) = \tv{u}$\, iff \,$V(\alpha,\neg f) = \tv{u}$. 
\item \label{If V(alpha,AND F) = tv{t} then V(alpha,f) = tv{t}} %6)
	If \,$V(\alpha,\AND F) = \tv{t}$\, then 
	for each $f$ in $F$, \,$V(\alpha,f) = \tv{t}$. 
\item \label{If V(alpha,f) = tv{t} then V(alpha,OR F) = tv{t}} %7)
	If \,$f \e F$\, and \,$V(\alpha,f) = \tv{t}$\, 
	then \,$V(\alpha,\OR F) = \tv{t}$. 
\item \label{If alpha isin {phi,pi,psi,beta} then V(alpha,f) isin {t,f,u}.} %8)
	If \,$\alpha \e \{\varphi,\pi,\psi,\theta,\theta',\beta\}$\, 
	then \,$V(\alpha,f) \e \{\tv{t},\tv{f},\tv{u}\}$. 
\item \label{If V(alpha,f) = a then alpha isin {psi',pi'}.} %9)
	If \,$V(\alpha,f) = \tv{a}$\, then \,$\alpha \e \{\psi',\pi'\}$. 
\end{compactenum}
\begin{compactenum}[1)] \raggedright
\addtocounter{enumi}{9}
\item \label{alpha is complete} %10)
	If \,$V(\alpha,f) = \tv{t}$\, then \,$\alpha \proves f$. (completeness)
\item \label{alpha is sound} %11)
	If \,$\alpha \e \{\varphi,\pi,\psi,\theta,\theta',\beta\}$\, and 
	\,$\alpha \proves f$\, then \,$V(\alpha,f) = \tv{t}$. (soundness)
\end{compactenum}
\end{Thm} %EndThm 6.2

%2345678901234567890123456789012345678901234567890123456789012345678901234567890
%Section 7
\section{Examples} 
\label{Section:Examples}
All the examples in my forthcoming book `Plausible Reasoning and Plausible Logic' (PRPL) 
are worked through in detail. 
Such workings are called evaluations and show how Plausible Logic (PL) deals with examples. 
The Principles in Section~\ref{Section:Principles} are shown to be satisfied by PL by construction, 
or results, or examples. 
The evaluations of such examples are proofs that the principle is satisfied; 
and so are omitted in this article. 

%2345678901234567890123456789012345678901234567890123456789012345678901234567890
%Section 8
\section{Conclusion} 
\label{Section:Conclusion}
This section highlights some of the important or new or unusual aspects covered in PRPL. 
We defined a rooted acyclic digraph (rad) and a new form of structural mathematical induction 
on rads. 
Seventeen principles of logics that do plausible reasoning are suggested and 
several important plausible reasoning examples are considered. 
There are 14 necessary principles and 3 desirable principles. 
A first-order logic, called Plausible Logic (PL), is defined that satisfies all but two of the 
desirable principles and reasons correctly with all the examples. 
As far as we are aware, this is the only such logic. 
PL has 8 reasoning algorithms which form a linear hierarchy. 
This is because, from a given plausible reasoning situation, there are different sensible conclusions. 
A formal proof or derivation in PL is a rad, which may be infinite. 

%2345678901234567890123456789012345678901234567890123456789012345678901234567890

\acks{Patrick Marchisella found 2 mistakes in an example, which eventually led to 
the definition of the proof algorithms $\theta$ and $\theta'$. Thanks Pat.}

\bibliographystyle{theapa}
\bibliography{Refs260309}

\end{document}